\documentclass[10pt,twocolumn,letterpaper]{article}

\usepackage{cvpr}
\usepackage{times}
\usepackage{epsfig}
\usepackage{graphicx}

\usepackage{amsmath,amsfonts,bm}

\def\eqref#1{equation~\ref{#1}}

\def\1{\bm{1}}

\DeclareMathAlphabet{\mathsfit}{\encodingdefault}{\sfdefault}{m}{sl}
\SetMathAlphabet{\mathsfit}{bold}{\encodingdefault}{\sfdefault}{bx}{n}

\usepackage{hyperref}
\usepackage{algorithm}
\usepackage[noend]{algorithmic}
\usepackage{url}
\usepackage{subcaption}
\usepackage{amsmath, amsthm, amssymb}
\usepackage[ansinew]{inputenc}
\usepackage{multirow}
\usepackage{booktabs}
\usepackage[flushleft]{threeparttable}
\usepackage{tabulary}
\usepackage{wrapfig}
\usepackage{lipsum}

\hypersetup{
colorlinks = true,
    urlcolor=blue,
}

\newcommand{\mmL}{\mathcal{L}}

\cvprfinalcopy 

\def\cvprPaperID{****} 

\begin{document}

\title{Towards Non-I.I.D. and Invisible Data with FedNAS: \\Federated Deep Learning via Neural Architecture Search}
\author{Chaoyang He\quad Murali Annavaram\quad Salman Avestimehr\\
University of Southern California\quad \\
{\tt\small chaoyang.he@usc.edu\quad annavara@usc.edu\quad avestime@usc.edu}
}

\maketitle

\begin{abstract}
Federated Learning (FL) has been proved to be an effective learning framework when data cannot be centralized due to privacy, communication costs, and regulatory restrictions. When training deep learning models under an FL setting, people employ the predefined model architecture discovered in the centralized environment. However, this predefined architecture may not be the optimal choice because it may not fit data with non-identical and independent distribution (non-IID).
Thus, we advocate automating federated learning (AutoFL) to improve model accuracy and reduce the manual design effort. We specifically study AutoFL via Neural Architecture Search (NAS), which can automate the design process. We propose a Federated NAS (FedNAS) algorithm to help scattered workers collaboratively searching for a better architecture with higher accuracy. We also build a system based on FedNAS. Our experiments on non-IID dataset show that the architecture searched by FedNAS can outperform the manually predefined architecture
\footnote{This paper was accepted to CVPR 2020 workshop on neural architecture search and beyond for representation learning. We released the source code at \url{https://fedml.ai}.}

\end{abstract}
\section{Introduction}
\begin{figure}[t]
\centering
\includegraphics[width=0.95\linewidth]{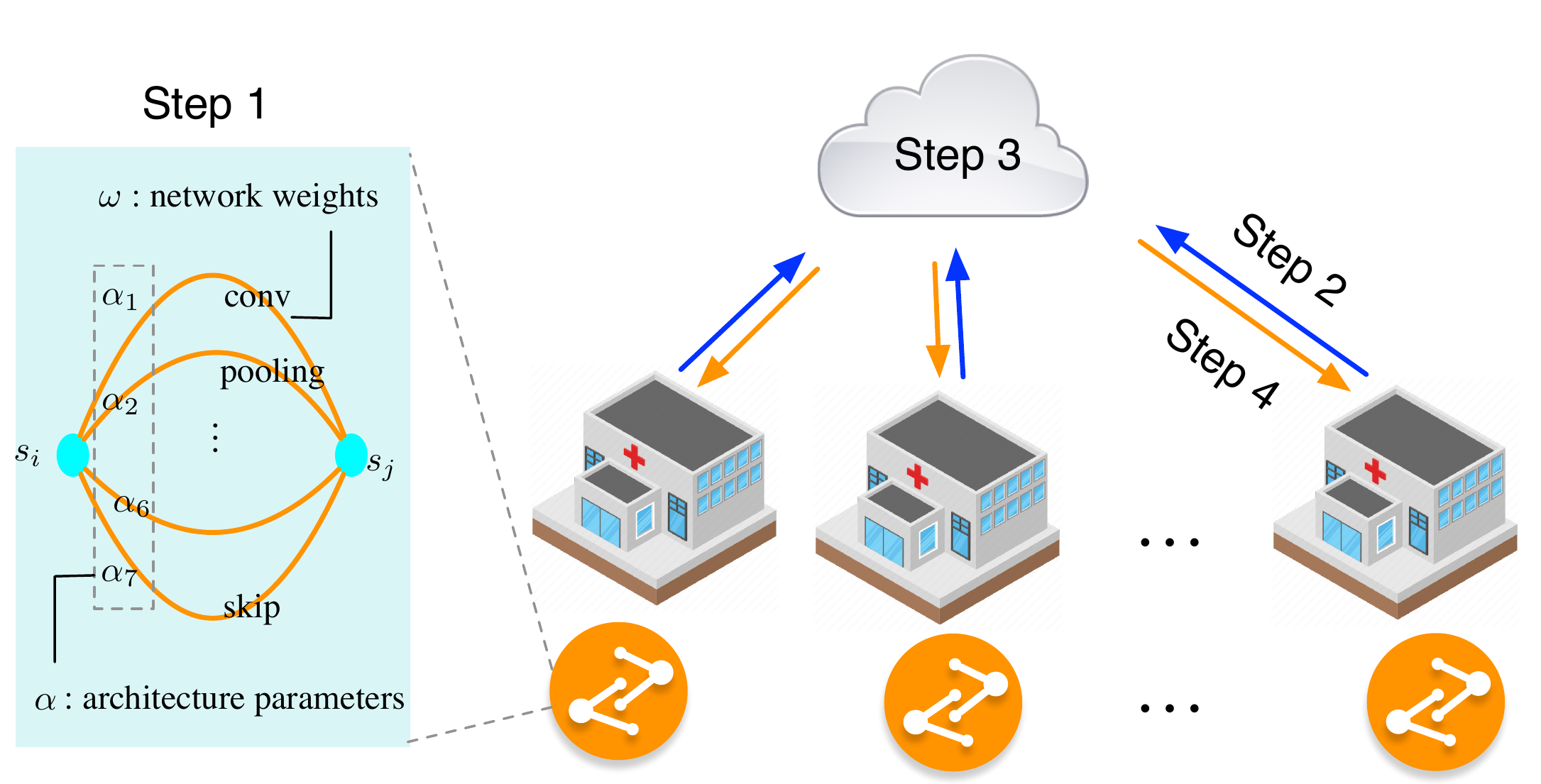}
\caption{Federated Neural Architecture Search (\textit{step 1}: search locally; \textit{step 2}: sending the gradient of $\alpha$ and $\omega$ to the server side; \textit{step 3}: merge gradient to get global $\alpha$ and $\omega$; \textit{step 4}: synchronize updated $\alpha$ and $\omega$ to each client.)}
\centering
\label{fig:overview}
\end{figure}

While CNN (Convolutional Neural Network) is capable of performing various computer vision tasks, it consumes a significant amount of data to achieve state-of-the-art performance. In the case of COVID-19, 
results from CT imaging, an effective method used to diagnose coronavirus-induced pneumonia, varies with the age and immunity status of the patient, as well as the disease stage at the time of scanning, underlying health conditions, and other drug interventions used at the time. 
Thus, significant amounts of CT image data from patients across hospitals around the world are essential for rapid and accurate diagnosis and treatment. However, centralizing such small and scattered heterogeneous data from various organizations is challenging due to privacy and confidentiality concerns, high communication and storage costs, protection of intellectual property, regulatory restrictions, and legal constraints. 
Consequently, a promising approach to solve this problem is Federated Learning (FL) \cite{mcmahan2016communication}, a decentralized learning framework which can train models without sharing the raw data with the collaborating nodes.
For example, in Figure \ref{fig:overview},  each hospital can each learn a CNN model without disclosing its own raw CT images.

However, as \cite{kairouz2019advances} points out, when training deep neural networks under the FL setting where the data is non-IID (non-identical and independent distribution), using the predefined model architecture may not be the optimal design choice. Since the data distribution is invisible to researchers, to find a better model architecture with higher accuracy, developers must design or choose multiple architectures, then tune hyperparameters remotely to fit the scattered data. This process is extremely expensive because attempting many rounds of training on edge devices results in a remarkably higher communication cost and on-device computational burden than the data center environment.

Due to these challenges, we advocate automating federated learning, which we term as AutoFL (Automated Federated Learning). In this work, we specifically study AutoFL via Neural Architecture Search (NAS), which automates the design process of the model architecture without inefficient manual attempts. In the centralized data environment, NAS significantly outperforms many manually designed architectures \cite{automl}.
However, it is still not clear whether NAS can really help boost the model performance in a heterogeneous distributed data environment like FL.
Therefore, we aim to answer a fundamental question:
\textit{can we design a collaborative NAS framework that helps improve model performance in the federated deep learning setting, in which the data distribution is non-identical and independent?} 

To answer this question, we consider the cross-organization scenario, in which each client in the network is a GPU-equipped edge server located in an organization (e.g., the hospital shown in Figure \ref{fig:overview}). We then propose Federated NAS (FedNAS) to search architectures among edge servers collaboratively. As the process shown in Figure \ref{fig:overview}, in FedNAS, we first utilize the gradient-based method MiLeNAS \cite{MiLeNAS} as local searcher on the local data of each worker. MiLeNAS can be easily distributed, and it is efficient in terms of search time. Formally, it formulates NAS as a mixed-level problem: $w = w - \eta_w \nabla_w \mmL_{\mathrm{tr}}(w, \alpha),
\alpha = \alpha - \eta_\alpha \left(\nabla_\alpha  \mmL_{\mathrm{tr}}(w, \alpha) +  \lambda \nabla_\alpha \mmL_{\mathrm{val}} (w, \alpha)\right)$, where $w$ represents the network weight and $\alpha$ determines the neural architecture.  $\mmL_{tr} (w, \alpha)$ and $\mmL_{val} (w, \alpha)$ denote the loss with respect to training data and validation data with $w$ and $\alpha$, respectively. After the local search, each worker then synchronizes weights $w$ and architecture $\alpha$ with other workers by weighted aggregation.

We design an AutoFL system based on the FedNAS to evaluate our idea. We construct a non-IID image dataset based on CIFAR10 and deploy our system in a distributed computing environment, which consists of 16 clients and 1 server. Our experiments show that Fed-NAS can search for a better architecture with a higher accuracy in only a few hours compared to \textit{FedAvg} \cite{mcmahan2016communication}, which purely utilizes the state-of-the-art manually designed architecture.

We also summarize our research directions that have the potential to further improve the efficiency and effectiveness of our proposed FedNAS.
\section{Related Works}
Recently, Neural Architecture Search (NAS) \cite{automl} has attracted widespread attention due to its advantages over manually designed models. There are three major NAS methods: evolutionary algorithms, reinforcement learning-based methods, and gradient-based methods \cite{MiLeNAS}. While in the Federated Learning (FL) domain \cite{mcmahan2016communication, he2019central}, using designed model architectures and optimizing by FedAvg \cite{mcmahan2016communication} is the main method to improve model performance. To our knowledge, NAS is rarely studied in FL setting.  Although \cite{kairouz2019advances} first proposed the concept of automating FL via NAS, the concrete method and details are never given. In our work, we propose a FedNAS algorithm and demonstrate its efficacy. Our AutoFL system design is relevant to \cite{bonawitz2019towards}. However, its system only supports manually designed architectures.
\section{Proposed Method}
\label{sec:overal}
\subsection{Problem Definition}
In federated learning setting, there are $K$ nodes in the network. Each node has a dataset $\mathcal{D}_{k}:=\left\{\left(x_{i}^{k}, y_{i}\right)\right\}_{i=1}^{N_{k}}$ which is non IID. When collaboratively training a deep neural networks (DNN) model with $K$ nodes, the objective function is defined as:
\begin{equation}
\min_{w} f(w,\underbrace{\alpha}_{fixed}) \stackrel{\text { def }}{=} \min_{w} \sum_{k=1}^{K} \frac{N_{k}}{N} \cdot \frac{1}{N_{k}} \sum_{i \in \mathcal{D}_{k}} \ell(x_{i}, y_{i}; w, \underbrace{\alpha}_{fixed} )
\end{equation}
where $w$ represents the network weight, $\alpha$ determines the neural architecture, and $\ell$ is the loss function of the DNN model. To minimize the objective function above, previous works choose a fixed model architecture $\alpha$ and then design variant optimization techniques to train the model $w$.

We propose to optimize the federated learning problem from a completely different angle, optimizing $w$ and $\alpha$ simultaneously. Formally, we can reformulate the objective function as:
\begin{equation}
\min_{w, \alpha} f(w,\alpha) \stackrel{\text { def }}{=} \min_{w, \alpha} \sum_{k=1}^{K} \frac{N_{k}}{N} \cdot \frac{1}{N_{k}} \sum_{i \in \mathcal{D}_{k}} \ell(x_{i}, y_{i}; w, \alpha )
\label{objective}
\end{equation}
In other words, for the non-IID dataset scattered across many workers, our goal is to search for an optimal architecture $\alpha$ and related model parameters $w$ to fit the dataset more effectively thus achieve better model performance. In this work, we consider searching for CNN architecture to improve the performance of the image classification task.

\subsection{Search Space}
\begin{figure}[h]
\centering
\includegraphics[width=1.0\linewidth]{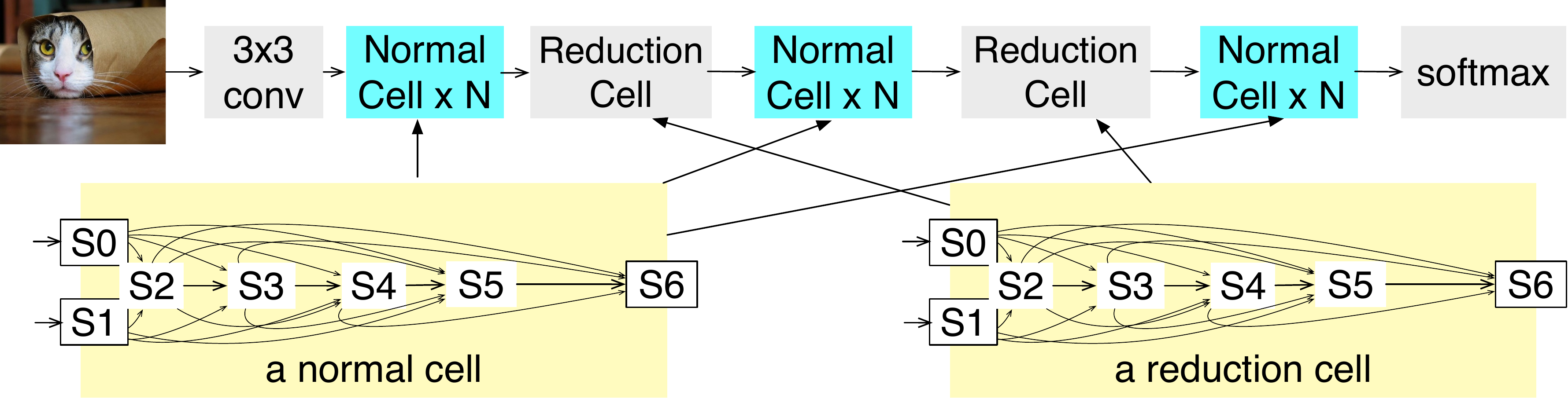}
\caption{Search Space}
\centering
\label{fig:searchspace}
\end{figure}
Normally, NAS includes three consecutive components: the search space definition, the search algorithm, and the performance estimation strategy \cite{automl}. 
Our search space follows the mixed-operation search space defined in DARTS \cite{liu2018darts} and MiLeNAS \cite{MiLeNAS}, where we search in two shared convolutional cells and then build it up as an entire model architecture (as shown in Figure \ref{fig:searchspace}). Inside the cell, to relax the categorical candidate operations between two nodes (e.g., convolution, max pooling, skip connection, zero) to a continuous search space, mixed operation using \text{softmax} over all possible operations is proposed:
\begin{equation}
\bar{o}^{(i,j)} (x) = \sum_{k=1}^{d} \underbrace{\frac{\exp(\alpha_k^{(i,j)})}{\sum_{k'=1}^{d} \exp(\alpha_{k'}^{(i,j)})}}_{p_k} o_k(x)
\label{equ: mixed operations}
\end{equation}
where the weight $p_k$ of the mixed operation $\bar{o}^{(i,j)} (x)$ for a pair of nodes $(i,j)$ is parameterized by a vector $\alpha^{i,j}$.
Thus, all architecture operation options inside a network (model) can be parameterized as $\alpha$. More details are introduced in Appendix \ref{sec:app_space}.

\subsection{Local Search}
Following the above-mentioned search space, each worker searches locally by utilizing the mixed-level optimization technique MiLeNAS \cite{MiLeNAS}:
\begin{equation}\label{eq:mixed_level2}
    \begin{aligned}
&w = w - \eta_w \nabla_w \mmL_{\mathrm{tr}}(w, \alpha)\\
  &\alpha = \alpha - \eta_\alpha \left(\nabla_\alpha  \mmL_{\mathrm{tr}}(w, \alpha) +  \lambda \nabla_\alpha \mmL_{\mathrm{val}} (w, \alpha)\right)
    \end{aligned}
\end{equation}
where $\mmL_{tr} (w, \alpha)$ and $\mmL_{val} (w, \alpha)$ denote the loss with respect to the local training data and validation data with $w$ and $\alpha$, respectively.

\subsection{Federated Neural Architecture Search}
We propose FedNAS, a distributed neural architecture search algorithm that aims at optimizing the objective function in Equation \ref{objective} under the FL setting. We introduce FedNAS corresponding to four steps in Figure \ref{fig:overview}: 1) The local searching process: each worker optimizes $\alpha$ and $w$ simultaneously using Eq. \ref{eq:mixed_level2} for several epochs; 2) All clients send their $\alpha$ and $w$ to the server; 3) The central server aggregates these gradients as follows:
\begin{equation}\label{eq:aggregation}
\begin{aligned}
    w_{t+1} \leftarrow \sum_{k=1}^{K} \frac{N_{k}}{N} w_{t+1}^{k}\\
    {\alpha}_{t+1} \leftarrow \sum_{k=1}^{K} \frac{N_{k}}{N} {\alpha}_{t+1}^{k}
\end{aligned}
\end{equation}
4) The server sends back the updated $\alpha$ and $w$ to clients, and each client updates its local $\alpha$ and $w$ accordingly, before running the next round of searching. This process is summarized in Algorithm \textcolor{red}{1}. 
After searching, an additional evaluation stage is conducted by using a traditional federated optimization method such as \textit{FedAvg} \cite{mcmahan2016communication}. Merging search and evaluation into one stage is one of future works (check Appendix \ref{sec:future_works} for more details).
\begin{algorithm}[htb]
    \caption{FedNAS Algorithm. }
    \begin{small}
        \begin{algorithmic}[1]
            \STATE \textbf{Initialization:} initialize $w_0$ and $\alpha_0$; $K$ clients are selected and indexed by $k$; $E$ is the number of local epochs; $T$ is the number of rounds.
            \STATE \textbf{Server executes:}
            \begin{ALC@g}
            \FOR{each round $t = 0, 1, 2, ..., T-1$}
            \FOR{each client $k$ \textbf{in parallel}}
            \STATE $w_{t+1}^{k}, {\alpha}_{t+1}^{k}  \leftarrow \text{ClientLocalSearch}(k, w_t, \alpha_t)$
            \ENDFOR
            \STATE $w_{t+1} \leftarrow \sum_{k=1}^{K} \frac{N_{k}}{N} w_{t+1}^{k}$
            \STATE ${\alpha}_{t+1} \leftarrow \sum_{k=1}^{K} \frac{N_{k}}{N} {\alpha}_{t+1}^{k}$
            \ENDFOR
            \end{ALC@g}
            \STATE 
            \STATE \textbf{ClientLocalSearch}($k$, $w$, $\alpha$): // \textit{Run on client $k$}
            \begin{ALC@g}
            \FOR{$e$ in epoch}
            \FOR{minibatch in training and validation data}
            \STATE Update $w = w - \eta_w \nabla_w \mmL_{\mathrm{tr}}(w, \alpha)$
            \STATE Update 
            \STATE $\alpha = \alpha - \eta_\alpha \left(\nabla_\alpha  \mmL_{\mathrm{tr}}(w, \alpha) +  \lambda \nabla_\alpha \mmL_{\mathrm{val}} (w, \alpha)\right)$
            \ENDFOR
            \ENDFOR
            \STATE return $w$, $\alpha$ to server
            \end{ALC@g}
        \end{algorithmic}
    \end{small}
\label{alg:mixed-level}
\end{algorithm}

\subsection{AutoFL System Design}
\label{sec:system}

\begin{figure}[h]
\centering
\includegraphics[width=\linewidth]{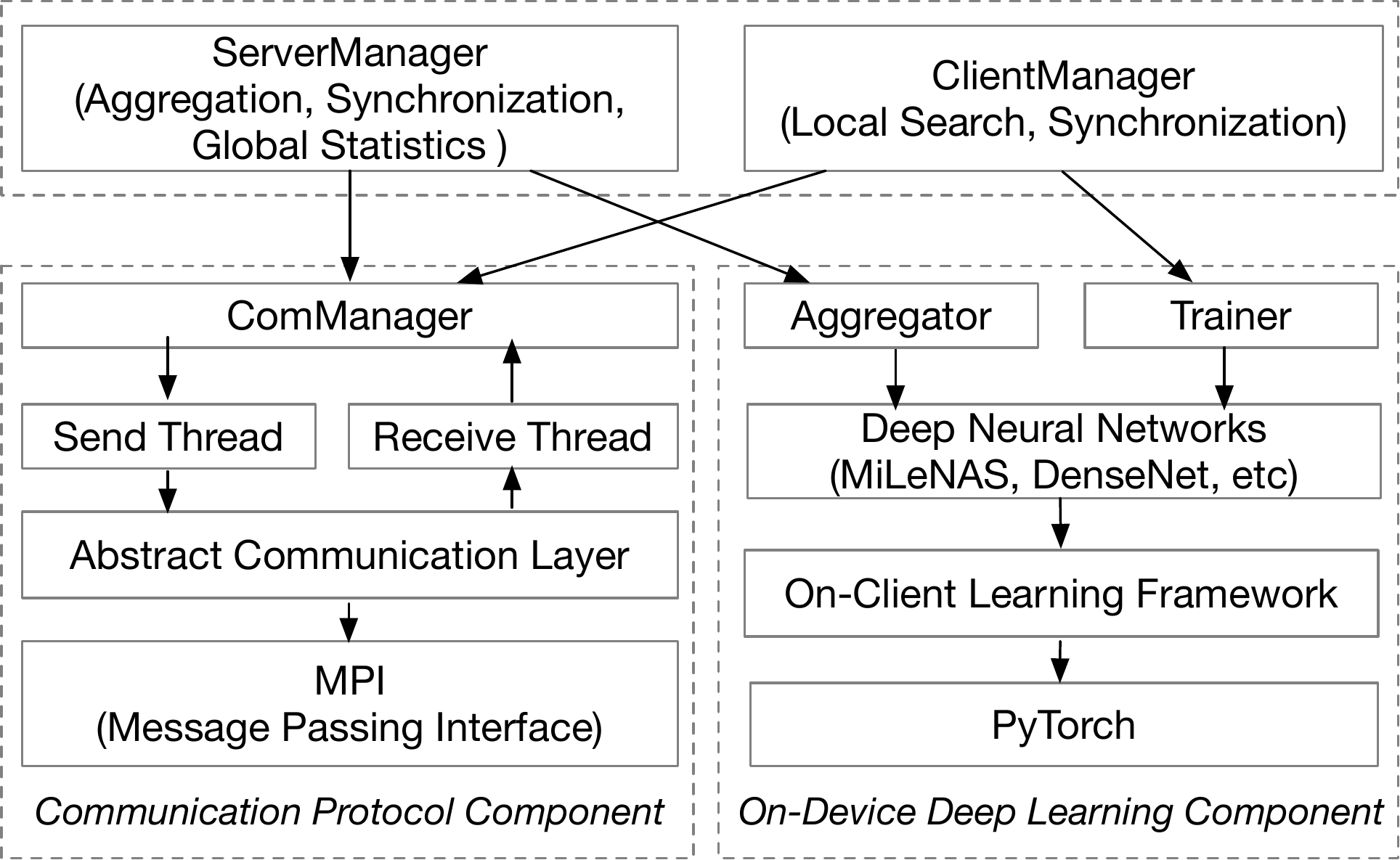}
\caption{Abstract System Architecture of AutoFL}
\centering
\label{fig:system}
\end{figure}

We design an AutoFL system using FedNAS based on \texttt{FedML} \cite{chaoyanghe2020fedml}, an open-source research library for federated learning. The system architecture is shown in Figure \ref{fig:system}. This design separates the communication and the model training into two core components shared by the server and clients. The first is the communication protocol component, which is responsible for low-level communication among the server and clients. The second is the on-device deep learning component, which is built based on the popular deep learning framework PyTorch. These two components are encapsulated as \textit{ComManager}, \textit{Trainer}, and \textit{Aggregator}, providing high-level APIs for the above layers. With the help of these APIs, in \textit{ClientManager}, the client can train or search for better architectures and then send its results to the server-side, while in \textit{ServerManager}, the server can aggregate and synchronize the model architecture and the model parameters with the client-side.

\section{Experiments and Results}
\subsection{Experimental Setup}
\textbf{Implementation and Deployment.} We set up our experiment in a distributed computing network equipped with GPUs. There are 17 nodes in total, one representing the server-side, and the other 16 nodes representing clients, which can be organizations in the real world (e.g., the hospitals). Each node is a physical server that has an NVIDIA RTX 2080Ti GPU card inside. We deployed the FedNAS system described in Appendix \ref{sec:system} on each node. Our code implementation is based on PyTorch 1.4.0, MPI4Py \footnote{\url{https://pypi.org/project/mpi4py/}} 3.0.3 , and Python 3.7.4. For simplicity, we assume all the clients join the training process for every communication round.

\textbf{Task and Dataset}. Our training task is image classification on the CIFAR10 dataset, which consists of 60000 32x32 color images in 10 classes, with 6000 images per class. We generate non-IID (non identical and independent distribution) dataset by splitting the 50000 training images into K clients in an unbalanced manner:  sampling $\mathbf{p}_{c} \sim \operatorname{Dir}_{J}(0.5)$ and allocating a $\mathbf{p}_{c, k}$ proportion of the training samples of class $c$ to local client $k$. The 10000 test images are used for a global test after the aggregation of each round. For different methods, we record the global test accuracy as the metric to compare model performance. Since the model performance is sensitive to the data distribution, we fix the non-IID dataset in all experiments for a fair comparison. The actual data distribution used for experiments can be found in Appendix \ref{sec:non-iid}.

\textbf{Model and Baseline}. We compare FedNAS with FedAvg. FedAvg runs on a manually designed architecture, DenseNet\cite{huang2017densely}, which extends ResNet \cite{he2016deep}, but higher performance and fewer model parameters than ResNet.


\subsection{Results on non-IID dataset}
\begin{figure}[hht]
\begin{subfigure}{.48\linewidth}
  \centering
    \includegraphics[width=\linewidth]{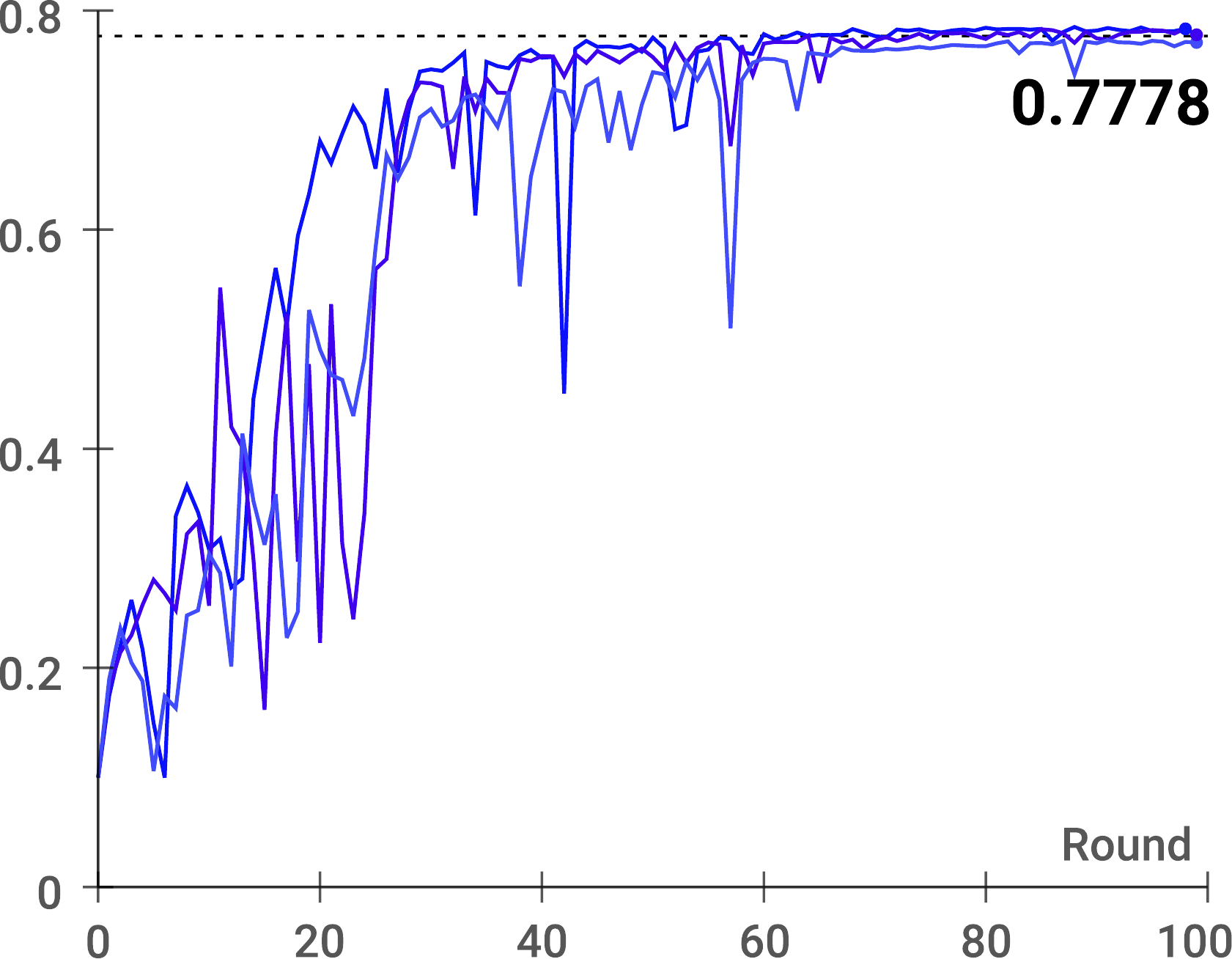}
    \caption{FedAvg on DenseNet}
    \label{fig:Searching_FedNAS_vs_FedAvg}
\end{subfigure}
\begin{subfigure}{.48\linewidth}
  \centering
    \includegraphics[width=\linewidth]{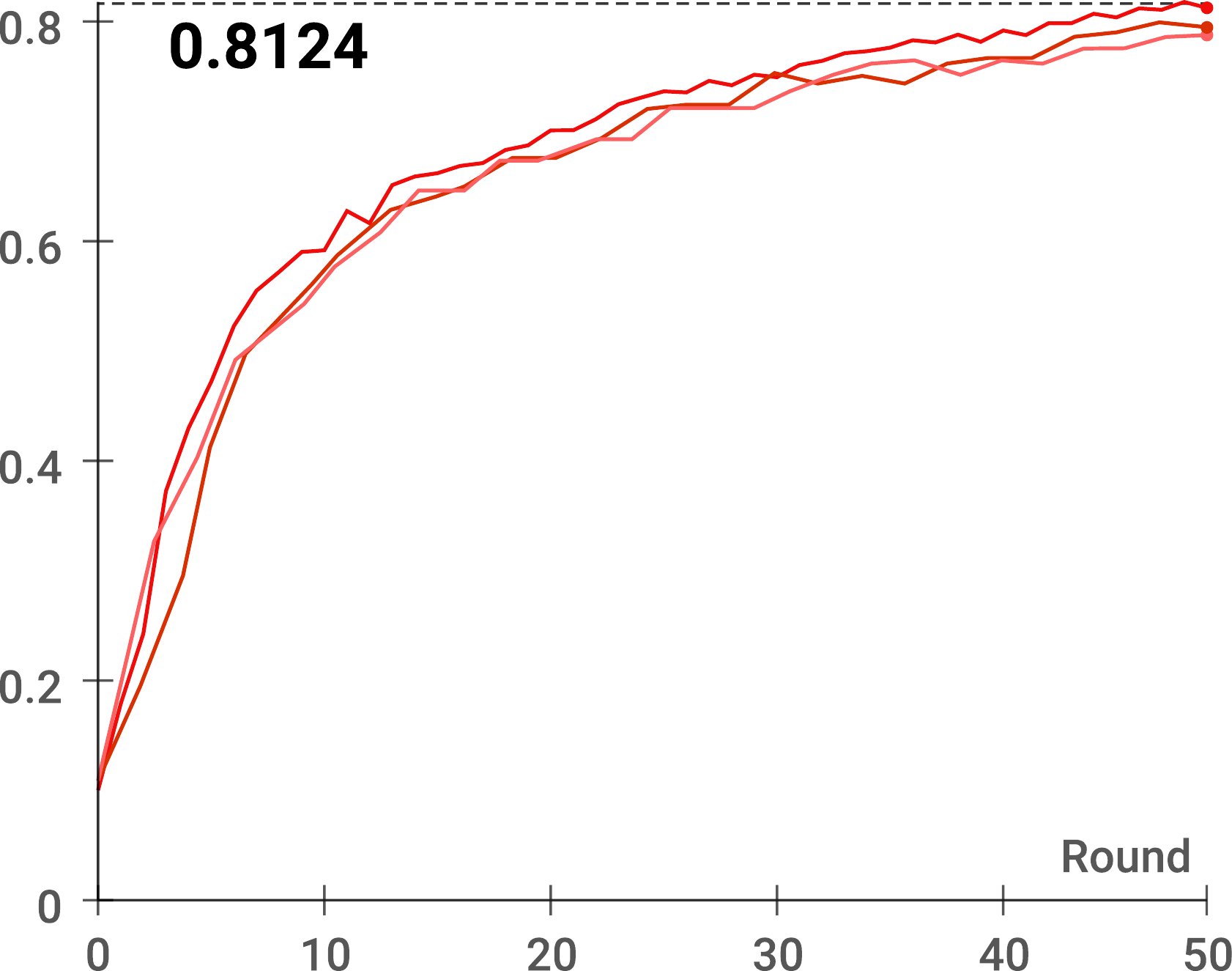}
    \caption{The search stage of FedNAS}
    \label{fig:acc_during_search}
\end{subfigure}
\caption{Test Accuracy on Non-IID Dataset (multiple runs)}
\label{fig:fig}
\end{figure}

Figure \ref{fig:Searching_FedNAS_vs_FedAvg} shows the results in a specific non-IID data distribution. For a fair comparison, results are obtained by fine-tuning hyperparameters of each method, and each method is run three times. 
Figure \ref{fig:acc_during_search} shows the test accuracy during the searching process. Interestingly, FedNAS can achieve higher accuracy than FedAvg during the searching process (81.24\% in Figure \ref{fig:acc_during_search}; 77.78\% in Figure \ref{fig:Searching_FedNAS_vs_FedAvg}). This further confirms the efficacy of FedNAS. We also evaluate the searched architecture under this data distribution. We found that each run of FedNAS can obtain a higher test accuracy than each run of FedAvg. On average, the  architecture searched by FedNAS obtains a test accuracy 4\% higher than FedAvg. 

Hyperparameters and visualization of the searched architecture can be found in Appendix \ref{sec:hpo} and \ref{sec:vis}, respectively.

\textbf{Remark}. \textit{We also run experiments on other distributions of non-IID datasets, in which FedNAS is also demonstrated to beat FedAvg, confirming that FedNAS searches for better architectures with a higher model performance.} 

\subsection{Evaluation of the Efficiency}
\begin{table}[hbt]
\resizebox{0.47\textwidth}{!}{
    \centering
    \begin{tabular}{c|c|c|c}
    \hline
      \multirow{2}{*}{\textbf{Method}} & \textbf{Search} & \textbf{Parameter} & \multirow{2}{*}{\textbf{Hyperparameter}} \\
       & \textbf{Time} & \textbf{Size} & \\
     
      \hline
       FedAvg  & \multirow{2}{*}{$>$ 3 days}  & \multirow{2}{*}{-} & rounds = 100\\
       (single)  &  &  & local epochs=20 \\
       \cline{1-3}
       FedAvg & \multirow{2}{*}{12 hours}  & \multirow{2}{*}{20.01M}& batch size=64 \\
      (distributed)  &   &  & \\
        \hline
       FedNAS  & \multirow{2}{*}{33 hours}  & \multirow{2}{*}{-} & rounds = 50\\
       (single)  &  &  & local epochs=5 \\
       \cline{1-3}
       \textbf{FedNAS} & \multirow{2}{*}{\textbf{$<$ 5 hours}}  & \multirow{2}{*}{\textbf{1.93M}}& batch size=64 \\
      \textbf{(distributed)}  &   &  & \\
       \hline
       
    \end{tabular}}
    \caption{Efficiency Comparison (16 RTX2080Ti GPUs as clients, and 1 RTX2080Ti as server)}
    \label{tab:searching time}
\end{table}

In order to more comprehensively reflect our distributed search overhead, we developed the single-process and distributed version of FedNAS and FedAvg. The single-process version simulates the algorithm by performing a client-by-client search on a single GPU card. As shown in Table \ref{tab:searching time}, compared to FedAvg and manually designed DenseNet, FedNAS can find better architecture with fewer parameters in less time. FedAvg spends more time because it requires more local epochs to converge.

\section{Future Works}
We introduce our future works in Appendix \ref{sec:future_works}.
\section{Conclusion}
We study automating federated learning (AutoFL) via Neural Architecture Search (NAS) by proposing a Federated NAS (FedNAS) algorithm that can help scatter workers collaboratively searching for a better architecture with higher accuracy. We build an AutoFL system based on FedNAS. Our experiments on the non-IID dataset show that the architecture searched by FedNAS can outperform FedAvg training on the manually designed architecture.

\small{
\bibliography{iclr2019_conference}
\bibliographystyle{ieee_fullname}
}

\appendix
\begin{table*}[hbt!]
    \centering
    \begin{tabular}{c|c|c|c|c|c|c|c|c|c|c|c}
    \hline
    \multirow{2}{*}{\textbf{Client ID}}  & \multicolumn{10}{c|}{\textbf{Numbers of samples in the classes}} &  \multirow{2}{*}{\textbf{Distribution}} \\
     \cline{2-11}
         & $c_0$ & $c_1$ & $c_2$ & $c_3$ & $c_4$ & $c_5$ & $c_6$ & $c_7$ & $c_8$ & $c_9$ &  \\
         \hline
       k=0 & 144 & 94 & \textbf{1561} & 133 & \textbf{1099} & \textbf{1466} & \textbf{0} & \textbf{0} & \textbf{0} & \textbf{0} & \begin{minipage}{0.15\textwidth}
      \includegraphics[width=\linewidth, height=0.15\textwidth]{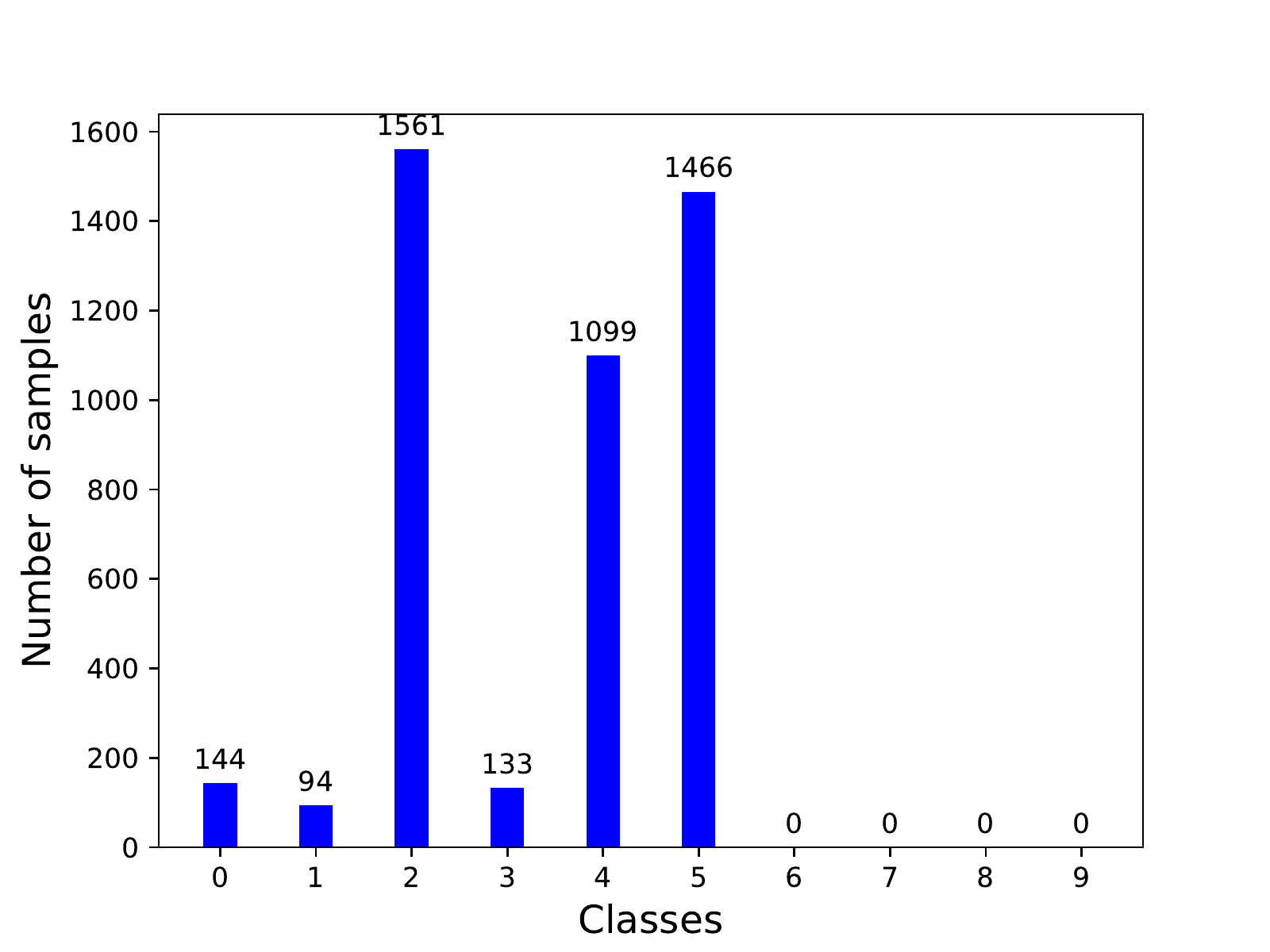}
    \end{minipage} \\
       \hline
       k=1 & 327 & 28 & 264 & 16 & 354  & 2 & 100 & 20 & 200 & 3 & \begin{minipage}{0.15\textwidth}
      \includegraphics[width=\linewidth, height=0.15\textwidth]{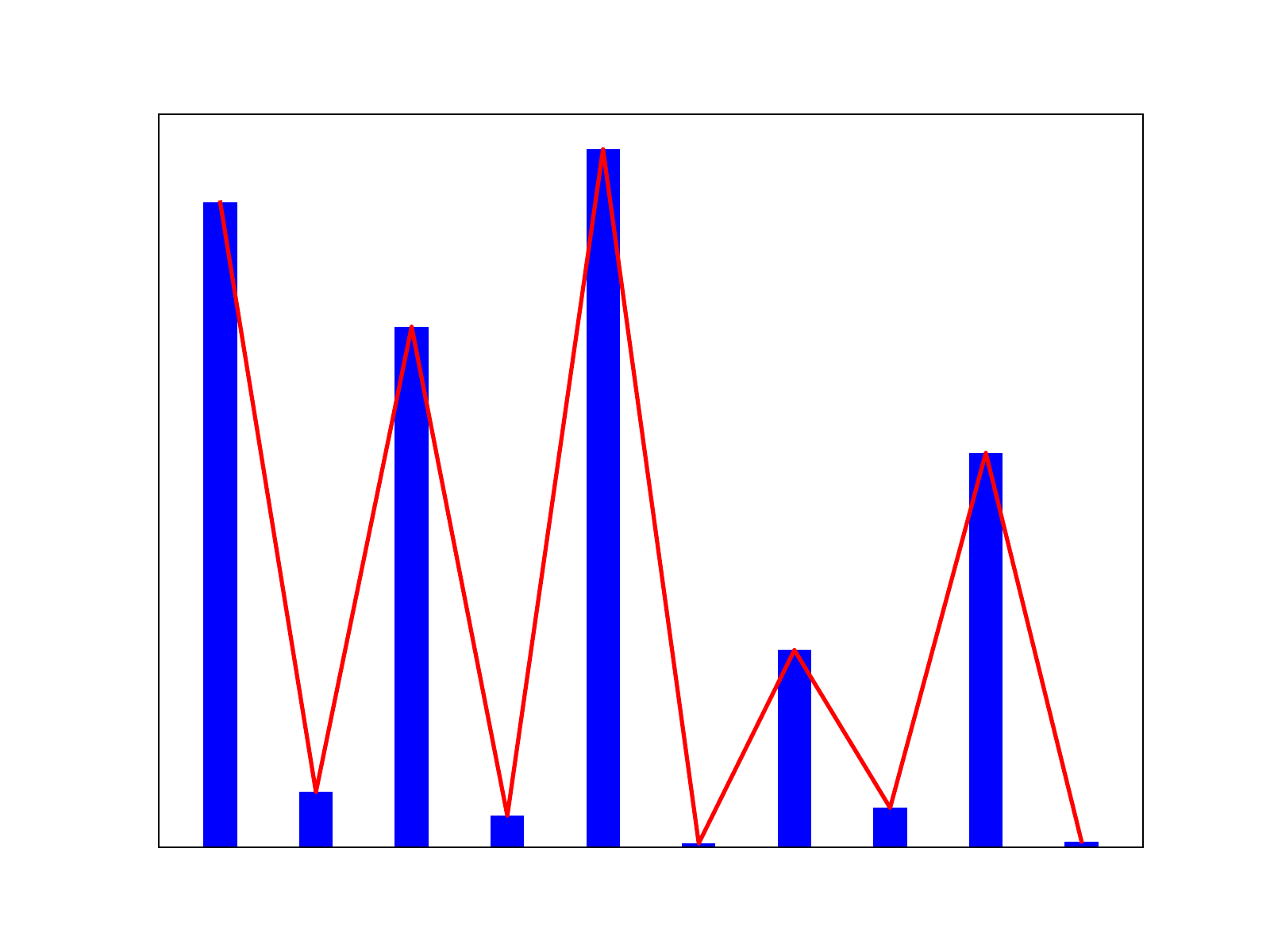}
    \end{minipage} \\
       \hline
       k=2 & 6 & 6 & 641 & 1 & 255 & 4 & 1 & 2 & 106 & \textbf{1723} & \begin{minipage}{0.15\textwidth}
      \includegraphics[width=\linewidth, height=0.15\textwidth]{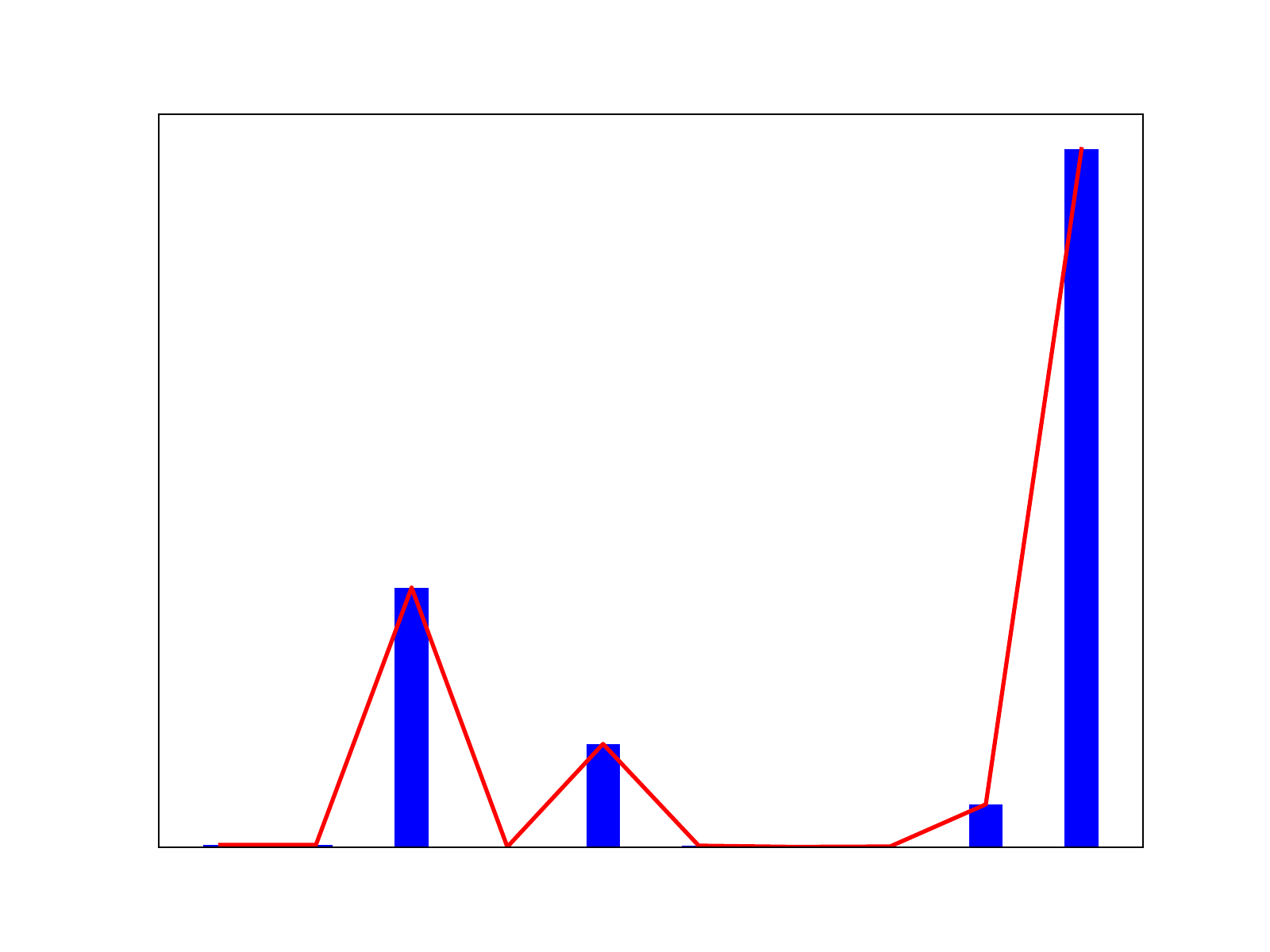}
    \end{minipage}  \\
       \hline
       k=3 & 176 & 792 & 100 & 28 & 76 & 508 & 991 & 416 & 215 & \textbf{0} & \begin{minipage}{0.15\textwidth}
      \includegraphics[width=\linewidth, height=0.15\textwidth]{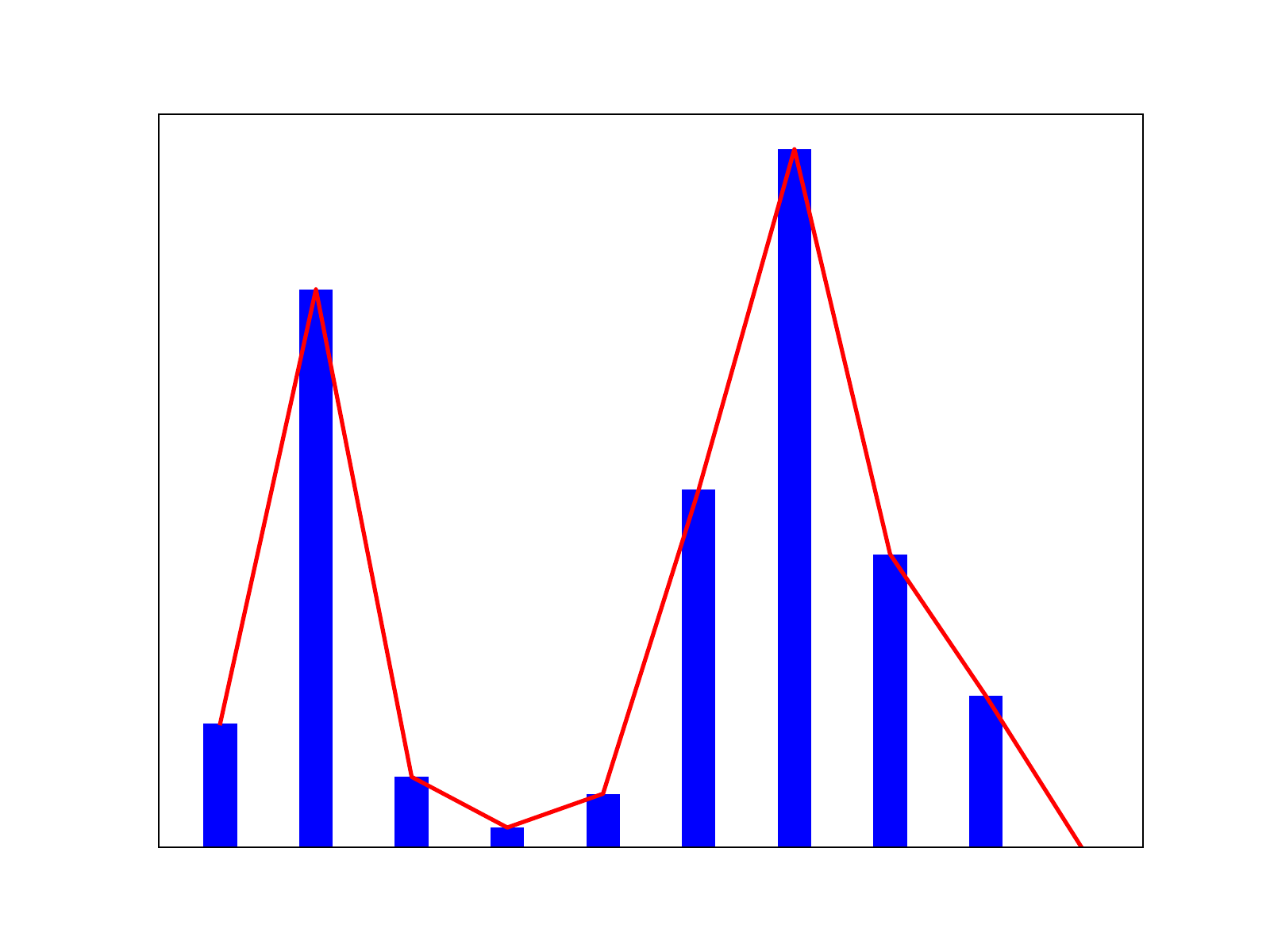}
    \end{minipage}  \\
       \hline
       k=4 & 84 & \textbf{1926} & 1 & 408 & 133 & 24 & 771 & \textbf{0} & \textbf{0} & \textbf{0} & \begin{minipage}{0.15\textwidth}
      \includegraphics[width=\linewidth, height=0.15\textwidth]{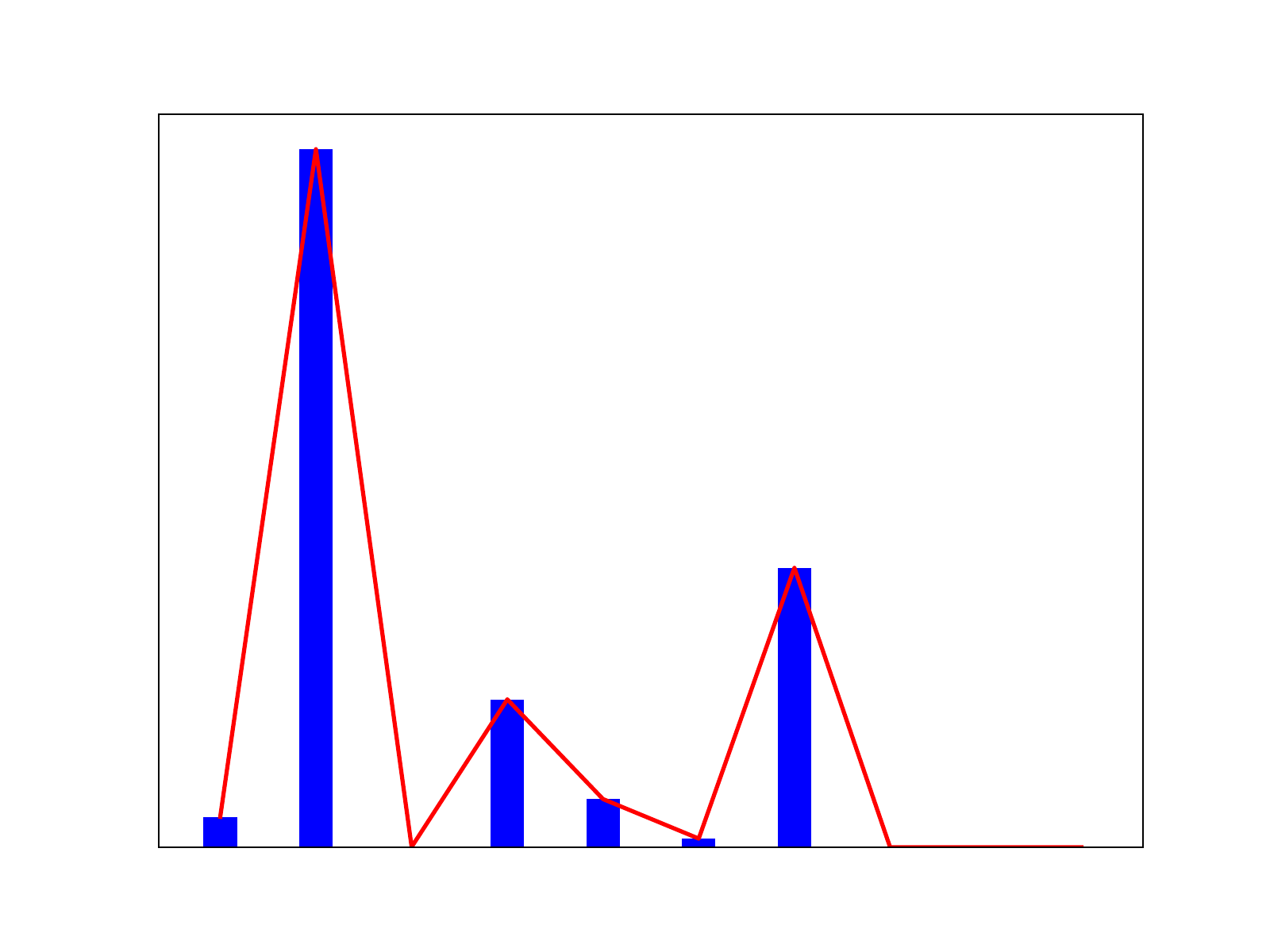}
    \end{minipage} \\
       \hline
       k=5 & 41 & 46 & 377 & 541 & 7 & 235 & 54 & \textbf{1687} & 666 & \textbf{0} & \begin{minipage}{0.15\textwidth}
      \includegraphics[width=\linewidth, height=0.15\textwidth]{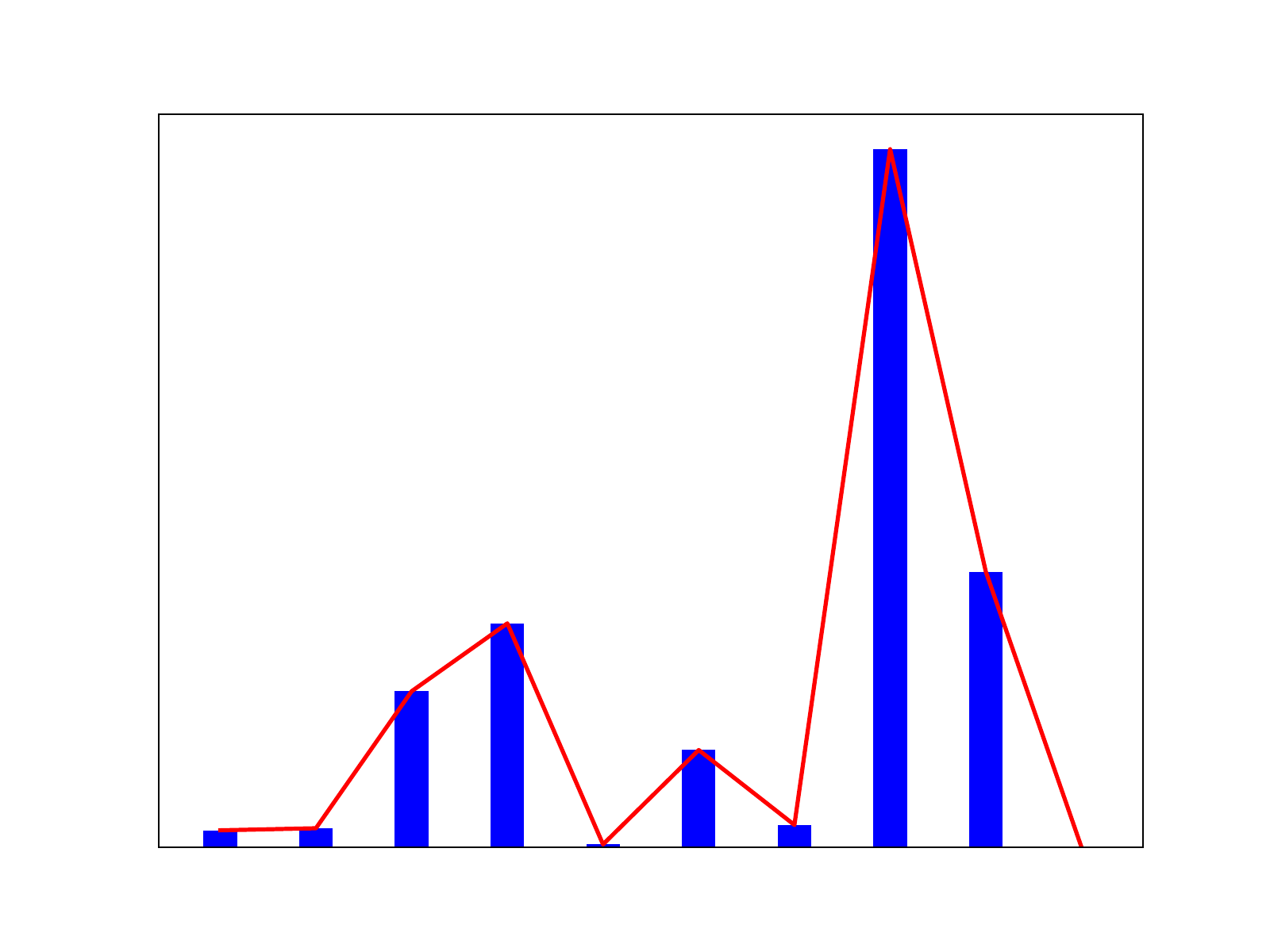}
    \end{minipage} \\
       \hline
       k=6 & 134 & 181 & 505 & 720 & 123 & 210 & 44 & 58 & 663 & 221 & \begin{minipage}{0.15\textwidth}
      \includegraphics[width=\linewidth, height=0.15\textwidth]{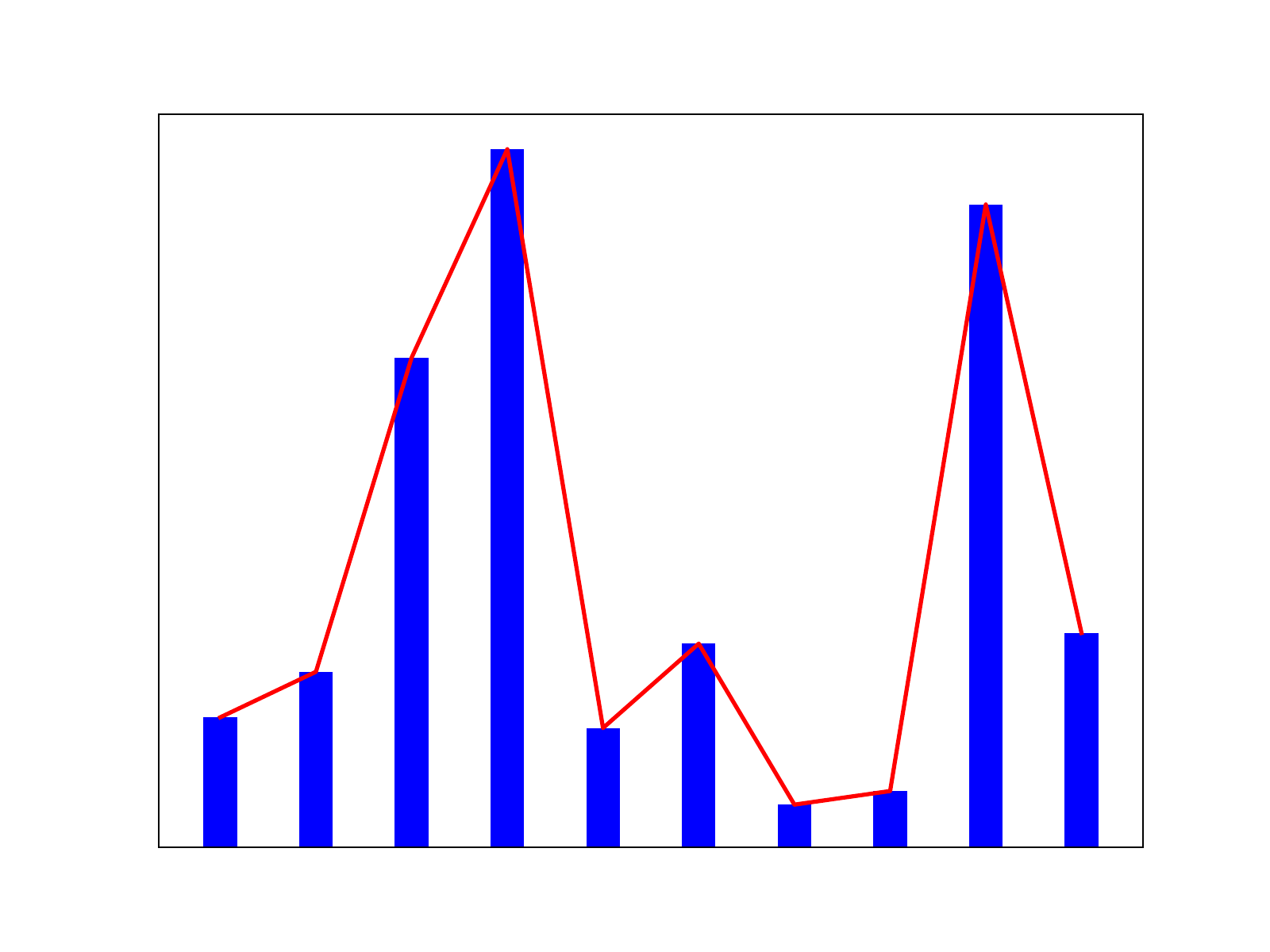}
    \end{minipage} \\
       \hline
       k=7 & 87 & 2 & 131 & \textbf{1325} & \textbf{1117} & 704 & \textbf{0} & \textbf{0} & \textbf{0} & \textbf{0}& \begin{minipage}{0.15\textwidth}
      \includegraphics[width=\linewidth, height=0.15\textwidth]{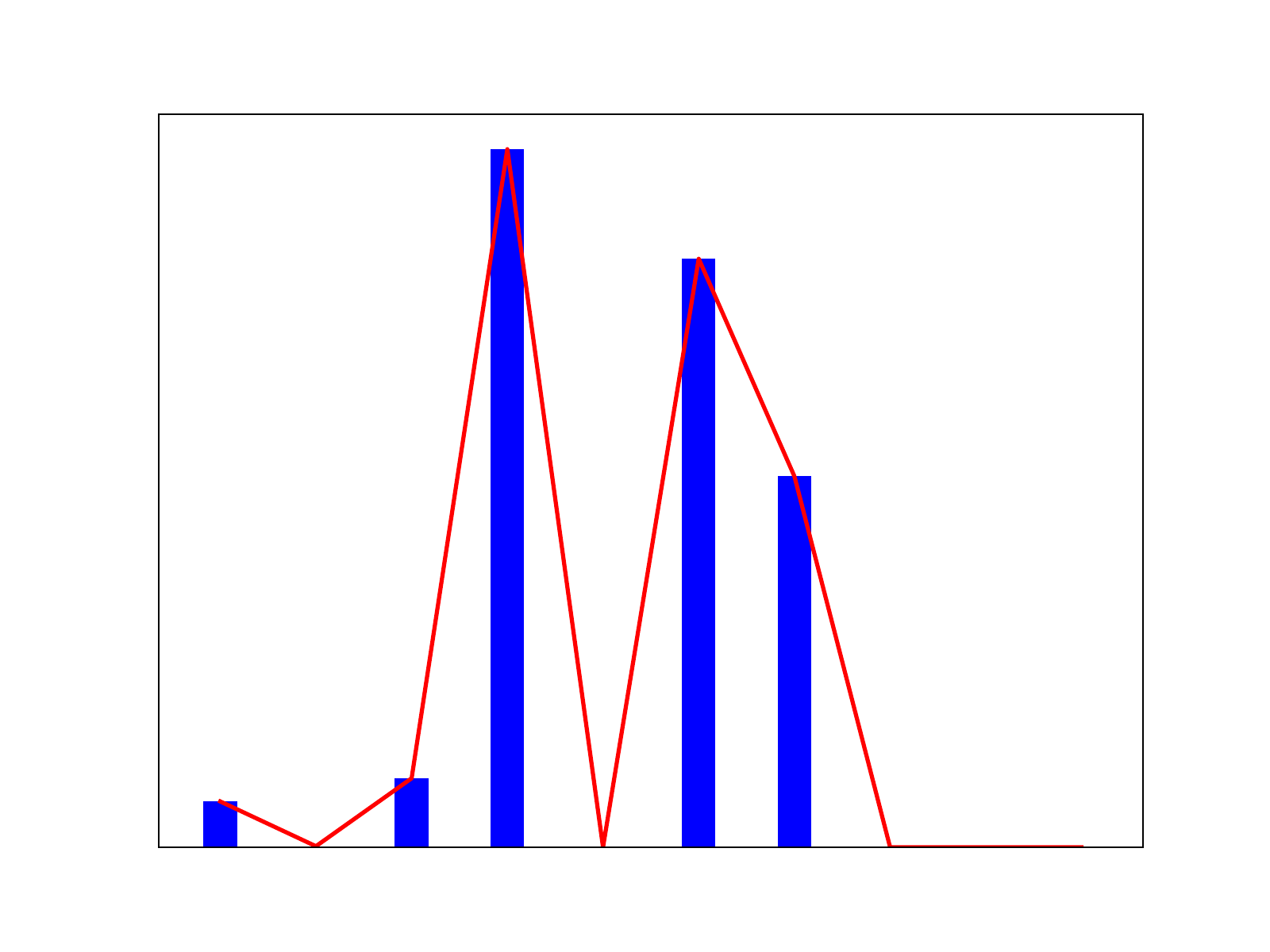}
    \end{minipage}  \\
       \hline
       k=8 & 178 & 101 & 5 & 32 & \textbf{1553} & 10 & 163 & 9 & 437 & 131 & \begin{minipage}{0.15\textwidth}
      \includegraphics[width=\linewidth, height=0.15\textwidth]{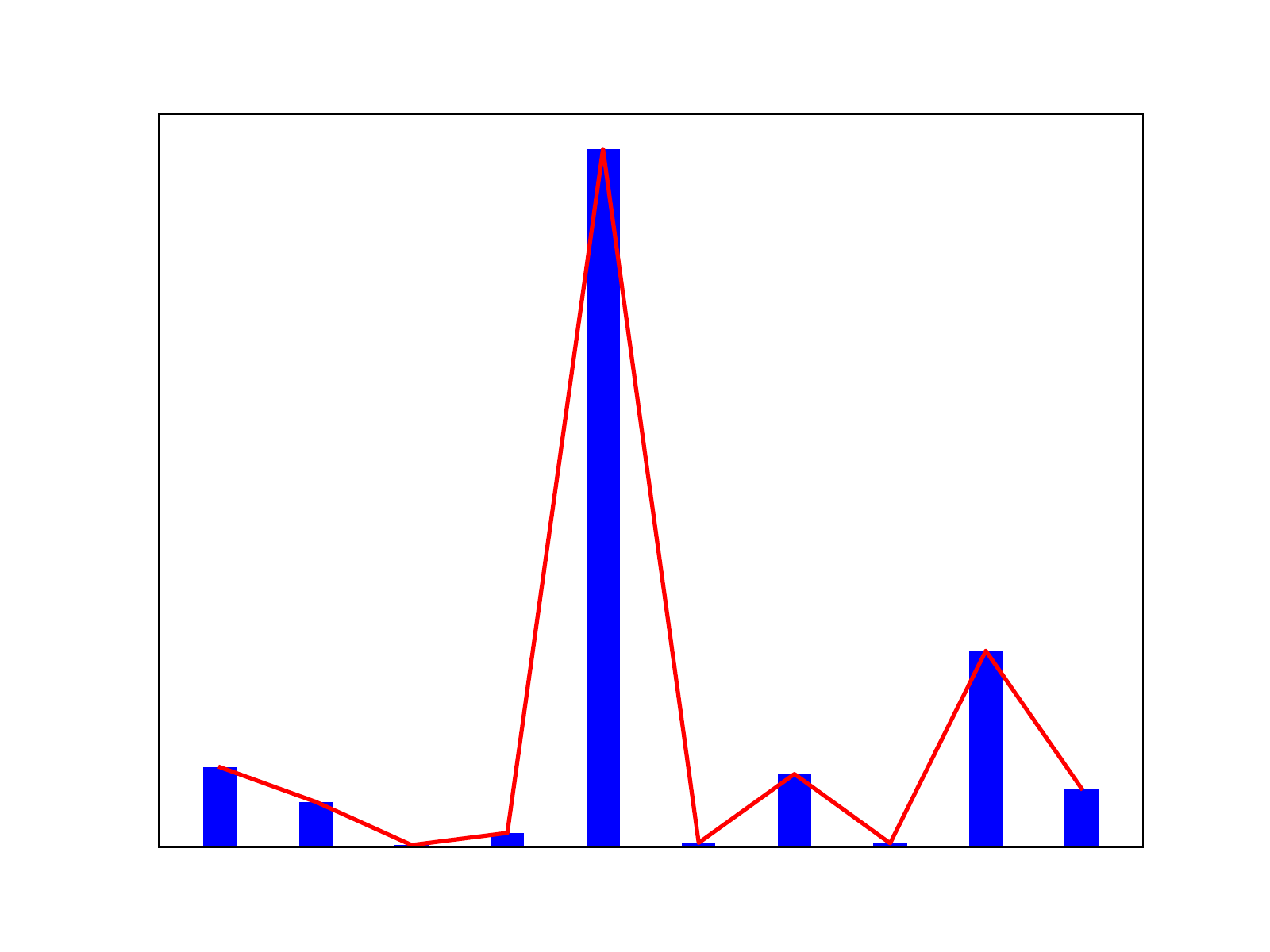}
    \end{minipage}  \\
       \hline
       k=9 & 94 & 125 & \textbf{0} & 147 & 287 & 100 & 23 & 217 & 608 & 279  & \begin{minipage}{0.15\textwidth}
      \includegraphics[width=\linewidth, height=0.15\textwidth]{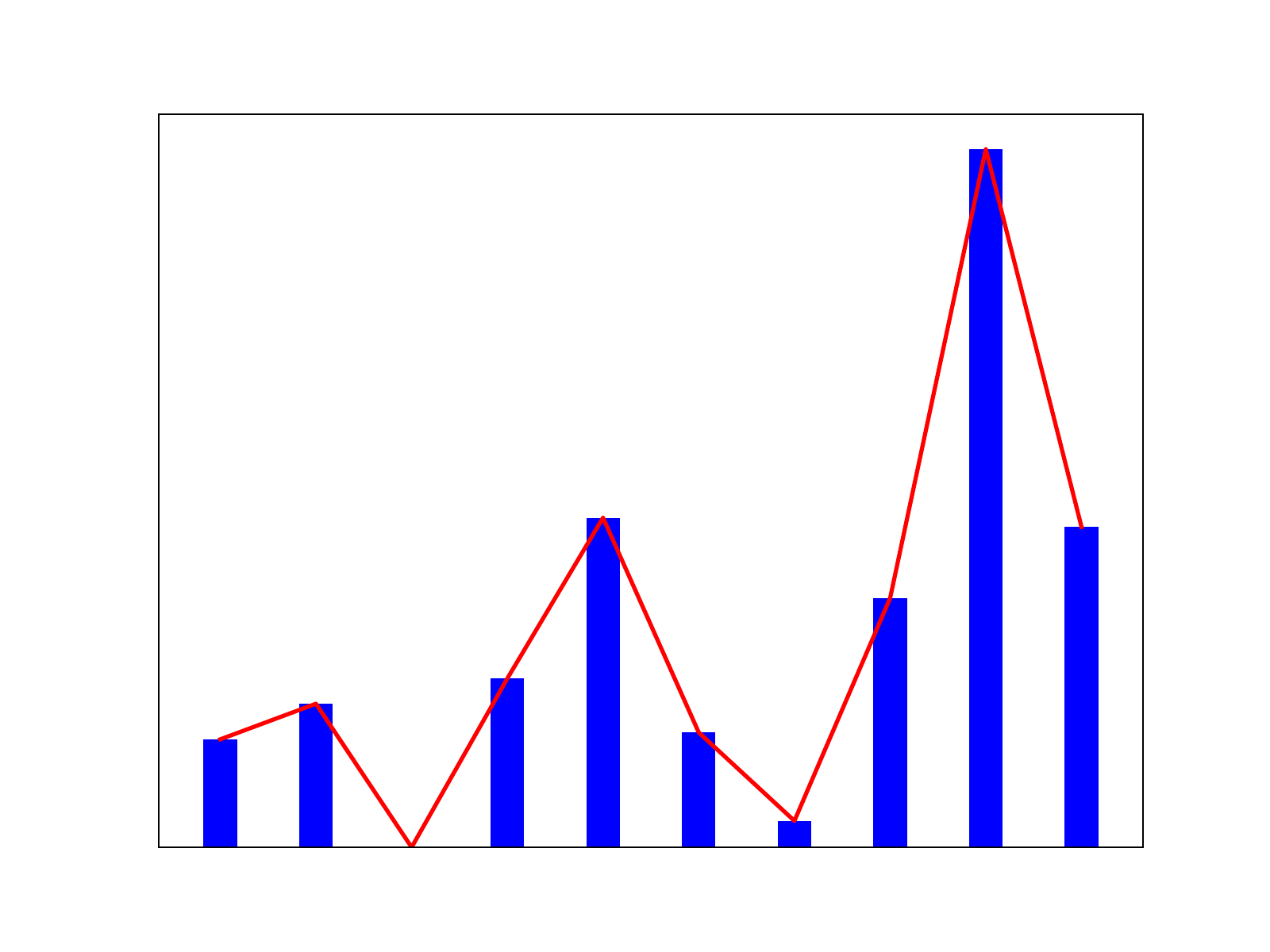}
    \end{minipage} \\
       \hline
       k=10 & 379 & 649 & 106 & 90 & 35 & 119 & 807 & 819 & 3 & 85 & \begin{minipage}{0.15\textwidth}
      \includegraphics[width=\linewidth, height=0.15\textwidth]{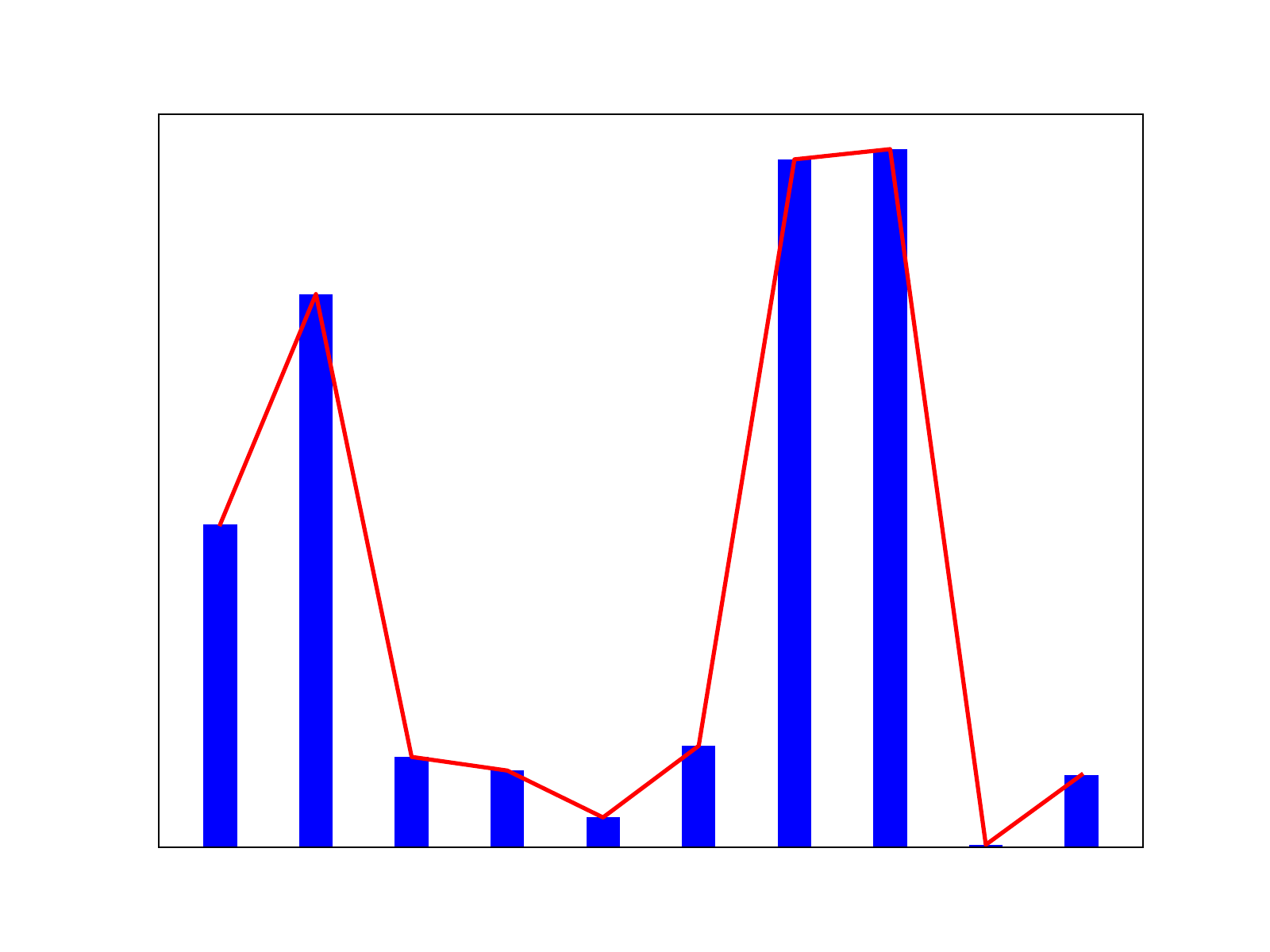}
    \end{minipage}  \\
       \hline
       k=11 & \textbf{1306} & 55 & 681 & 227 & 202 & 34 & \textbf{0} & 648 & \textbf{0} & \textbf{0} & \begin{minipage}{0.15\textwidth}
      \includegraphics[width=\linewidth, height=0.15\textwidth]{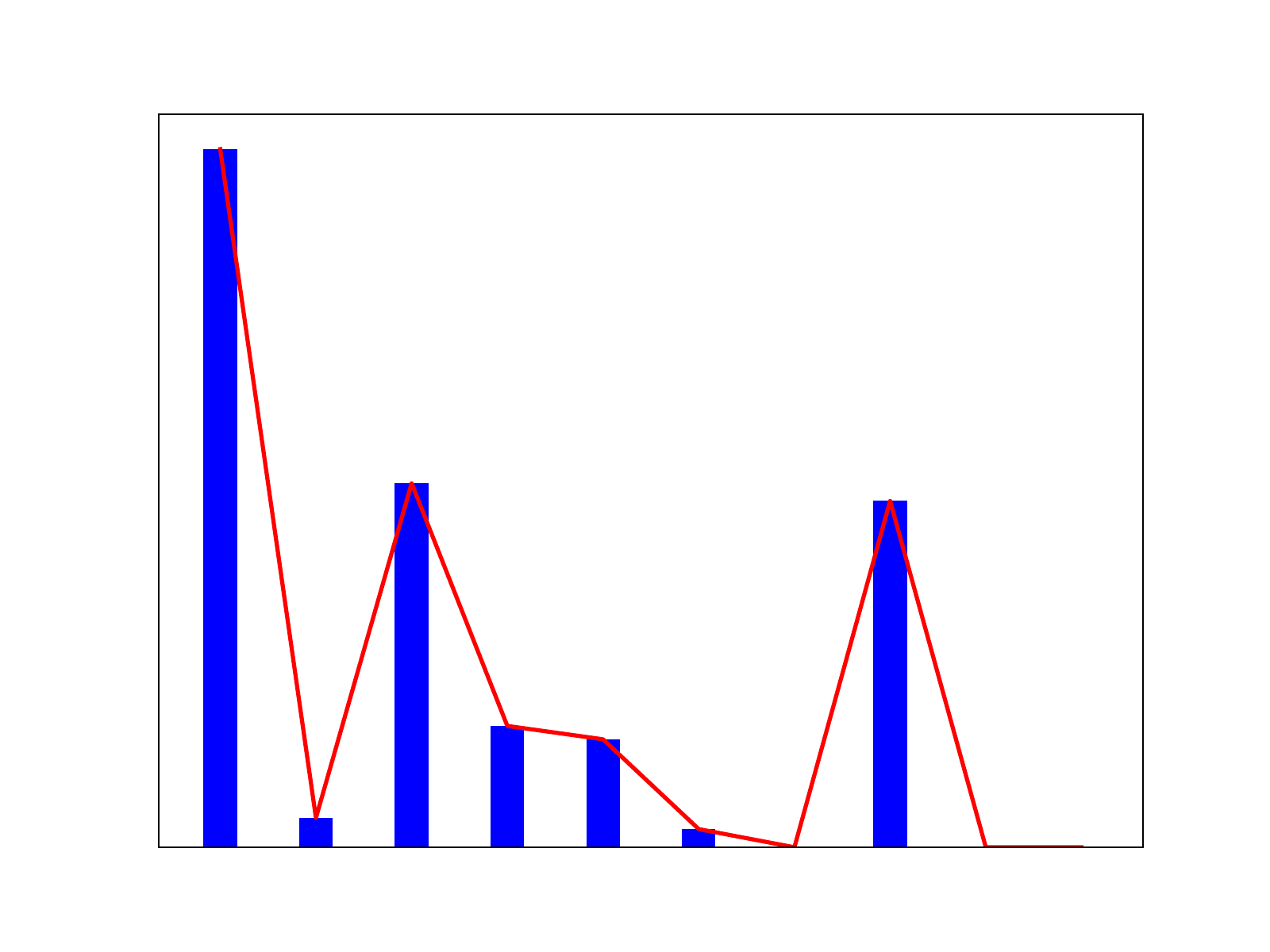}
    \end{minipage}  \\
       \hline
       k=12 & \textbf{1045}  & 13 & 53 & 6 & 77 & 70 & 482 & 7 & 761 & 494 & \begin{minipage}{0.15\textwidth}
      \includegraphics[width=\linewidth, height=0.15\textwidth]{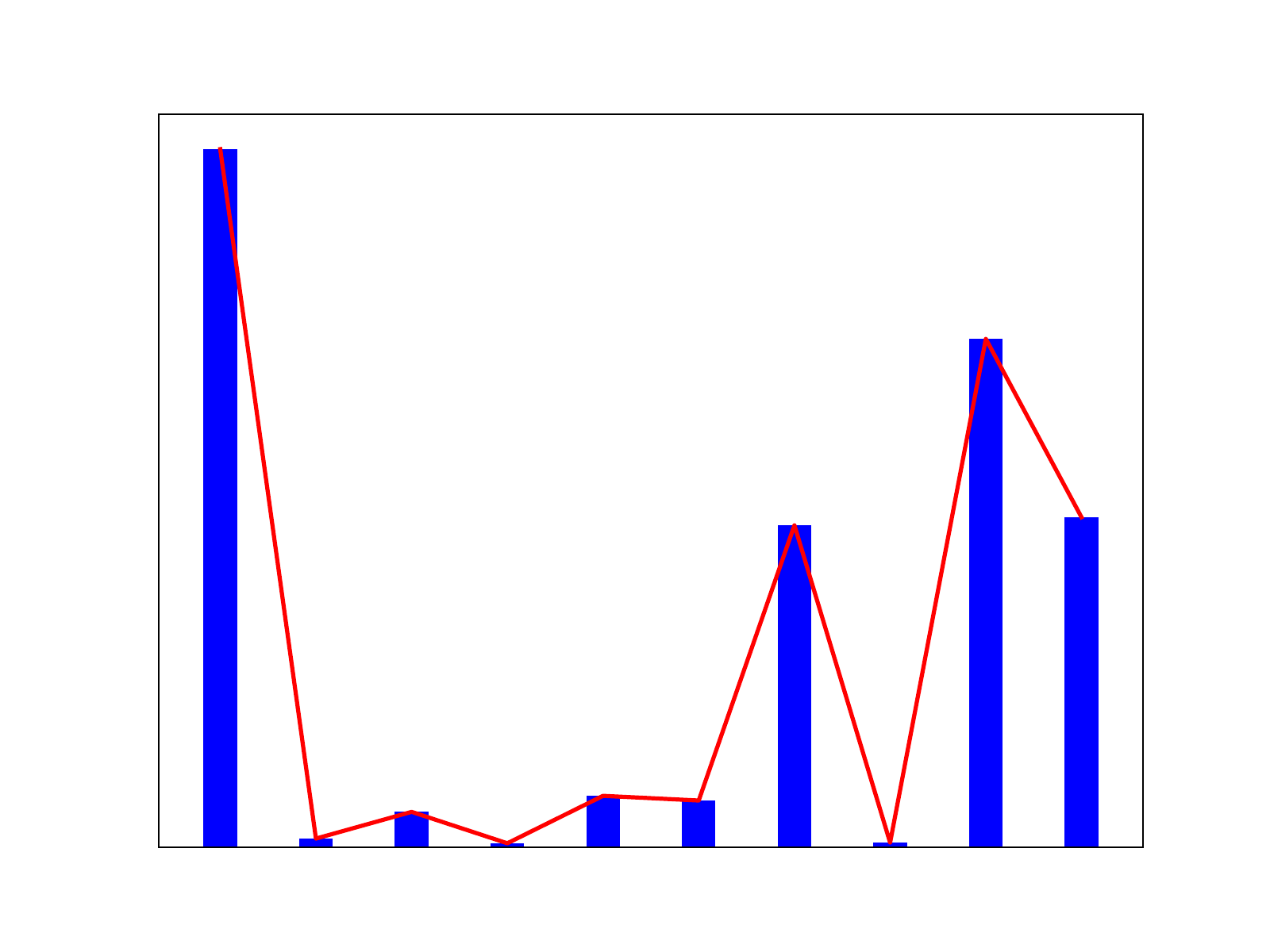}
    \end{minipage}  \\
       \hline
       k=13 & 731 & 883 & 15 & 161 & 387 & 552 & 4 & \textbf{1051} & \textbf{0} & \textbf{0} & \begin{minipage}{0.15\textwidth}
      \includegraphics[width=\linewidth, height=0.15\textwidth]{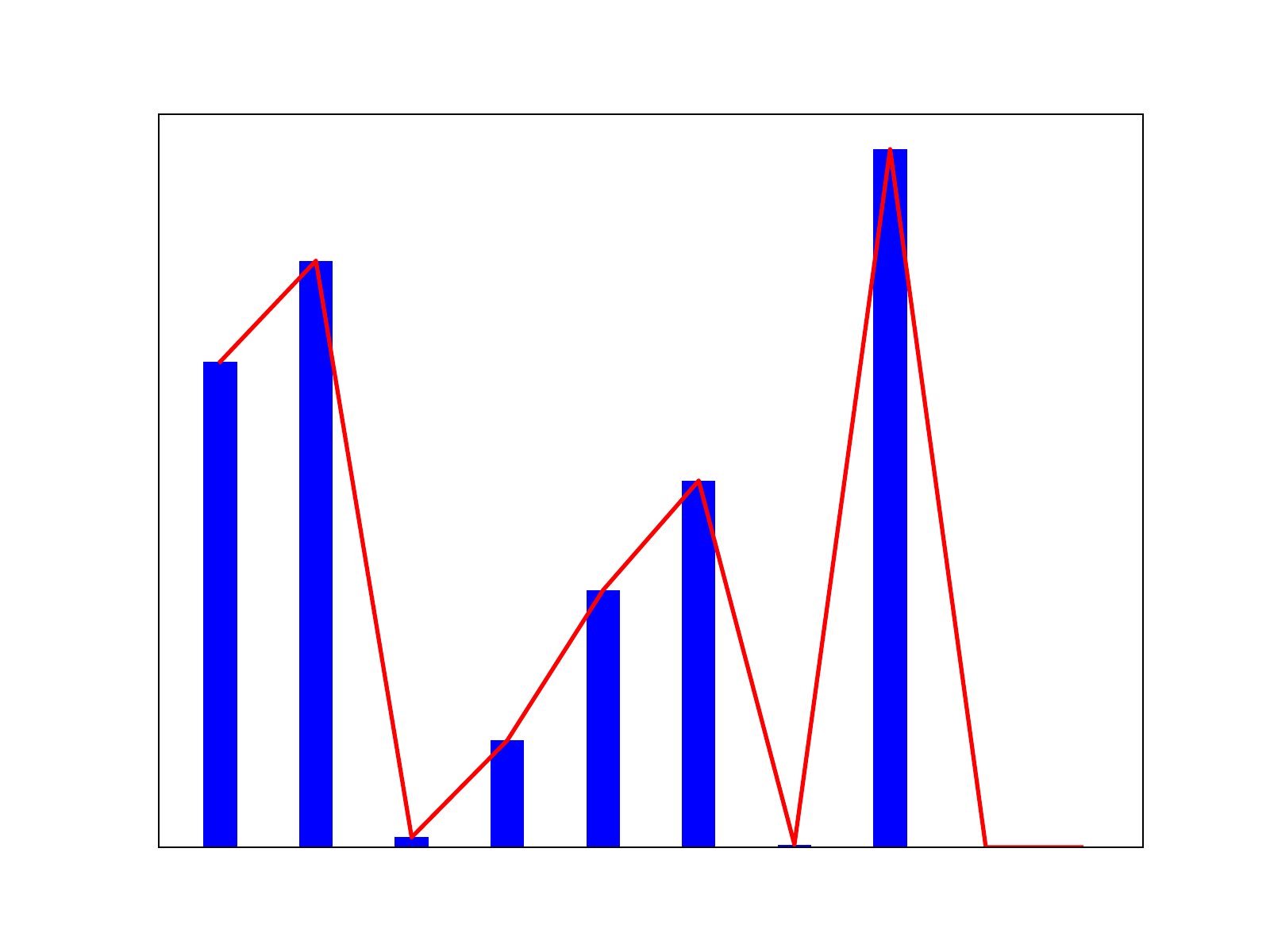}
    \end{minipage}  \\
       \hline
       k=14 & 4 & 97 & 467 & 899 & \textbf{0} & 407  & 50 & 64 & \textbf{1098} & 797 & \begin{minipage}{0.15\textwidth}
      \includegraphics[width=\linewidth, height=0.15\textwidth]{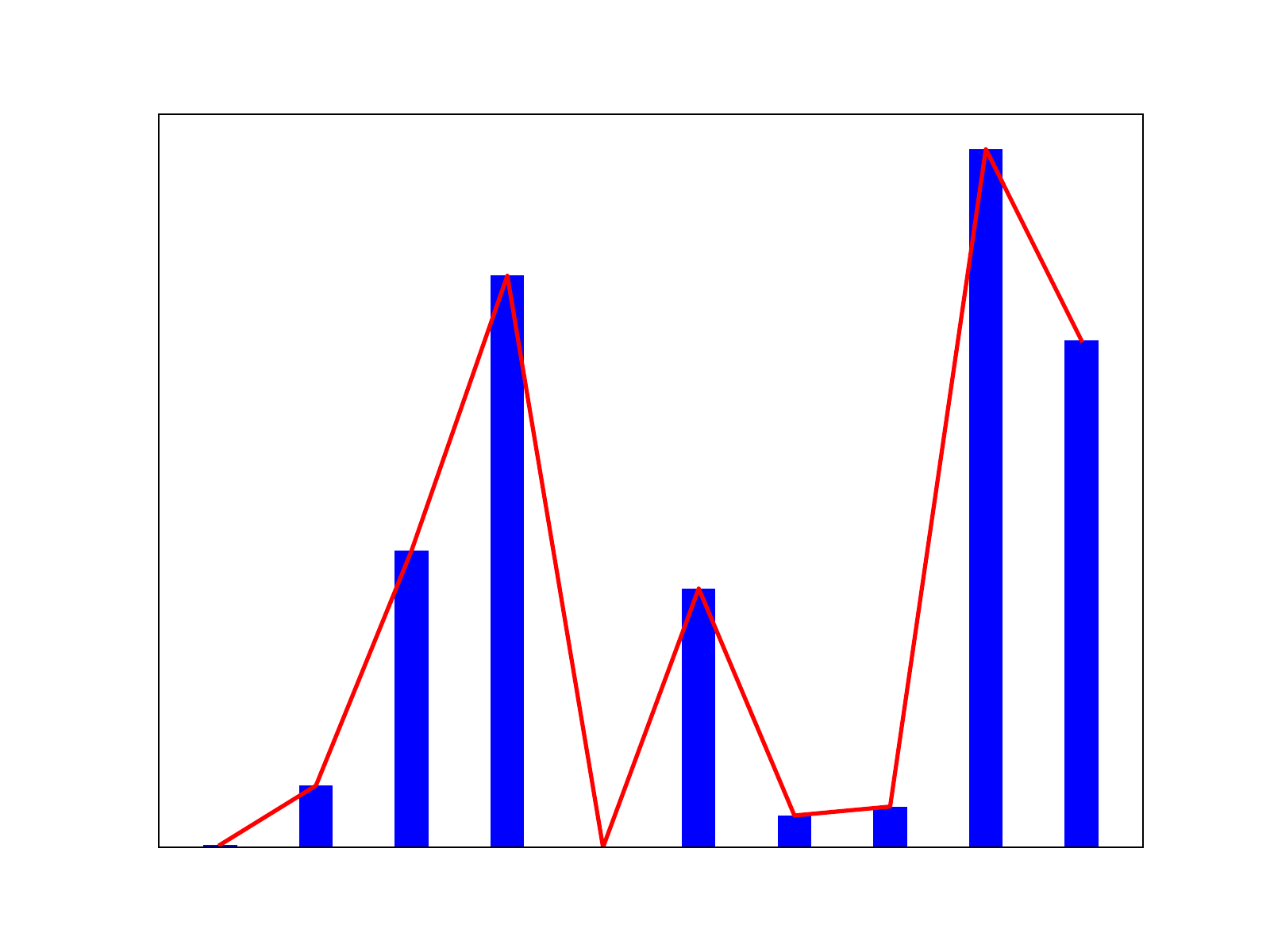}
    \end{minipage} \\
       \hline
       k=15 & 264 & 2 & 93 & 266 & 412  & 142 & 806 & 2 & 243 & \textbf{1267} & \begin{minipage}{0.15\textwidth}
      \includegraphics[width=\linewidth, height=0.15\textwidth]{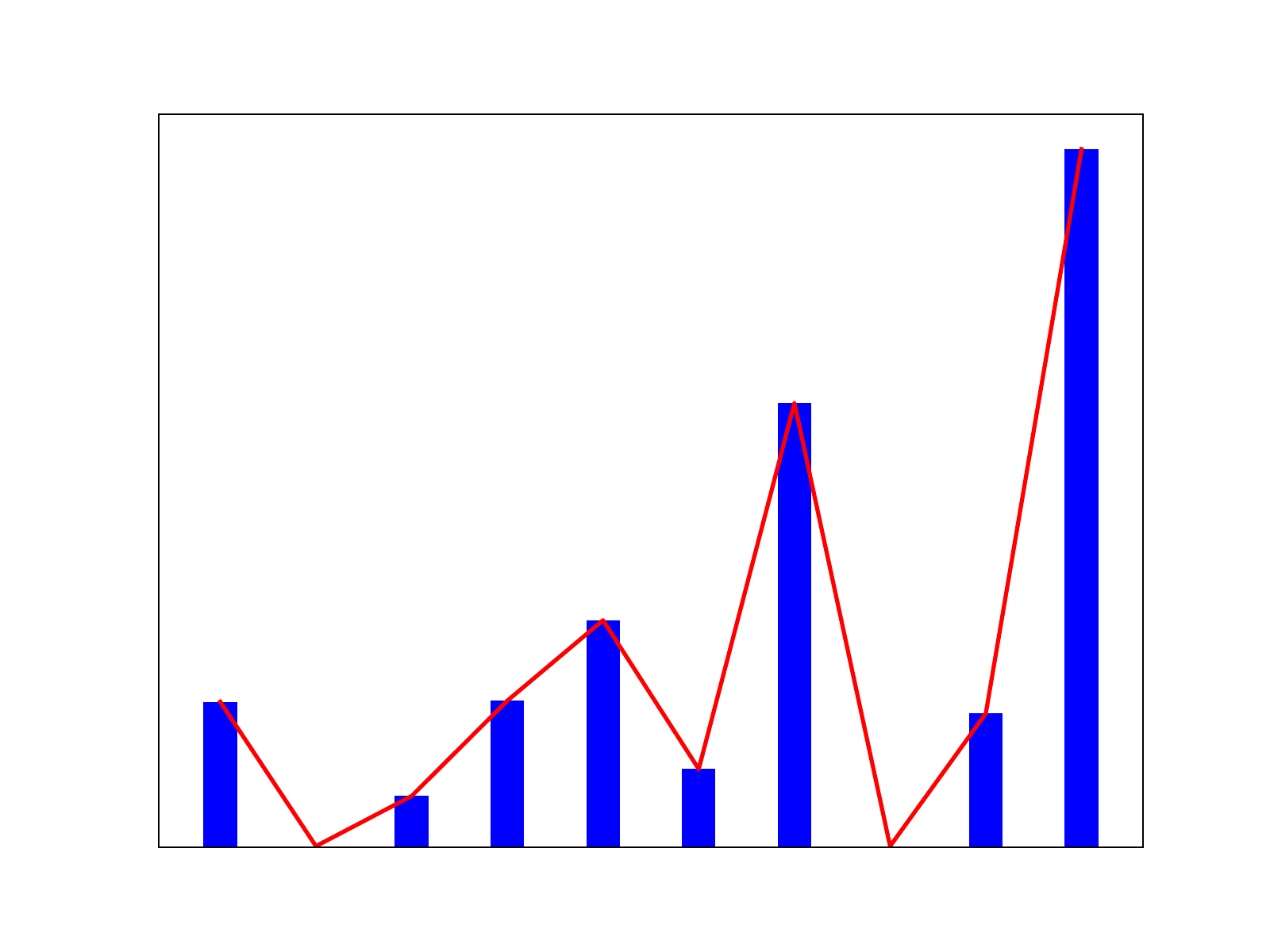}
    \end{minipage} \\
       \hline
    \end{tabular}
    \caption{The actual heterogeneous data distribution (non-IID) used in experiments}
    \label{tab:non-iid-distribution}
\end{table*}

\section{More Experiment Details}

\subsection{Details of the Search Space Definition}
\label{sec:app_space}
We adopt the following 7 operations in our CIFAR-10 experiments: 3 $\times$ 3 and 5 $\times$ 5 separable convolutions, 3 $\times$ 3 and $5 \times 5$ dilated separable convolutions, 3 $\times$ 3 max pooling, 3 $\times$ 3 average pooling, identity, and zero.

The network is formed by stacking convolutional cells multiple times. Cell $k$ takes the outputs of cell $k-2$ and cell $k-1$ as its input. Each cell contains seven nodes: two input nodes, one output node, and the other four intermediate nodes inside the cell. The input of the first intermediate node is set equal to two input nodes, and the other intermediate nodes take all previous intermediate nodes' output as input. The output node concatenates all intermediate nodes' output depth-wise. There are two types of cells: the normal cell and the reduction cell. The reduction cell is designed to reduce the spatial resolution of feature maps, located at 1/3 and 2/3 of the total depth of the network. Architecture parameters determine the discrete operation value between two nodes. All normal cells and all reduction cells share the same architecture parameters $\boldsymbol{\alpha}_{\text {n}}$ and $\boldsymbol{\alpha}_{\text {r}}$, respectively. By this definition, our method alternatively optimizes architecture parameters ($\boldsymbol{\alpha}_{\text {n}}$, $\boldsymbol{\alpha}_{\text {r}}$) and model weight parameters $\boldsymbol{w}$. 

\subsection{Details of the heterogeneous distribution on each client (non-IID)}
\label{sec:non-iid}
Table \ref{tab:non-iid-distribution} shows the actual data distribution used in our experiments. We can see that the sample number of each class in each worker is highly unbalanced. Some classes in a worker even have no samples, and some classes take up most of the proportion (highlighted in the table).

\subsection{Hyperparameter Setting} 
\label{sec:hpo}
We report important well-tuned hyperparameters used in our experiments. FedNAS searches 50 communication rounds using 5 local searching epochs, with a batch size of 64. For FedAvg, DenseNet201 is used for training, with 100 communication rounds, 20 local epochs, a learning rate of 0.08, and a batch size of 64. Both methods use the same data augmentation techniques that are used in image classification, such as random crops, flips, and normalization. More details and other parameter settings can be found in our source code.

\subsection{Visualization of the Search Architecture}
\label{sec:vis}
\begin{figure}[hbt!]
        \centering 
      \includegraphics[width=\linewidth]{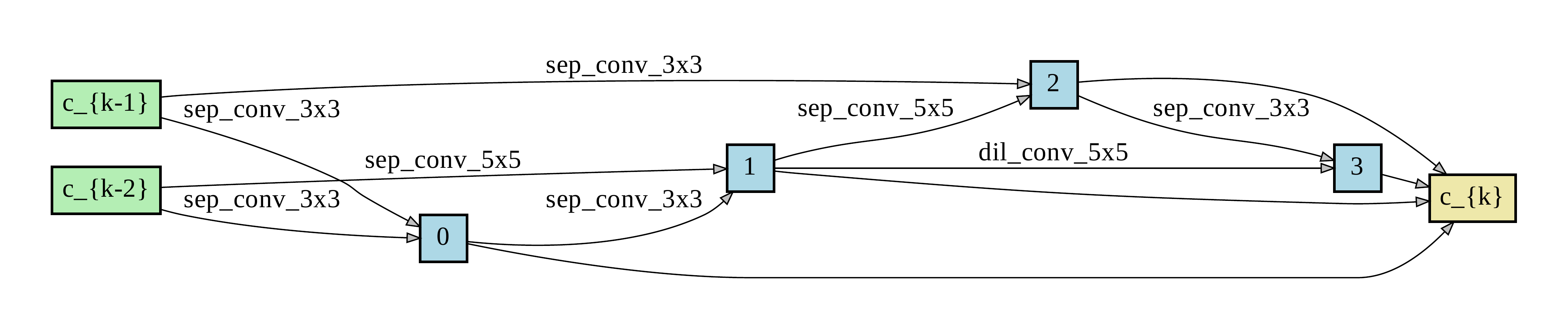}
    \caption{Normal Cell Architecture}
    \label{fig:normal}
\end{figure}
\begin{figure}[hbt!]
     \centering  
    \includegraphics[width=\linewidth]{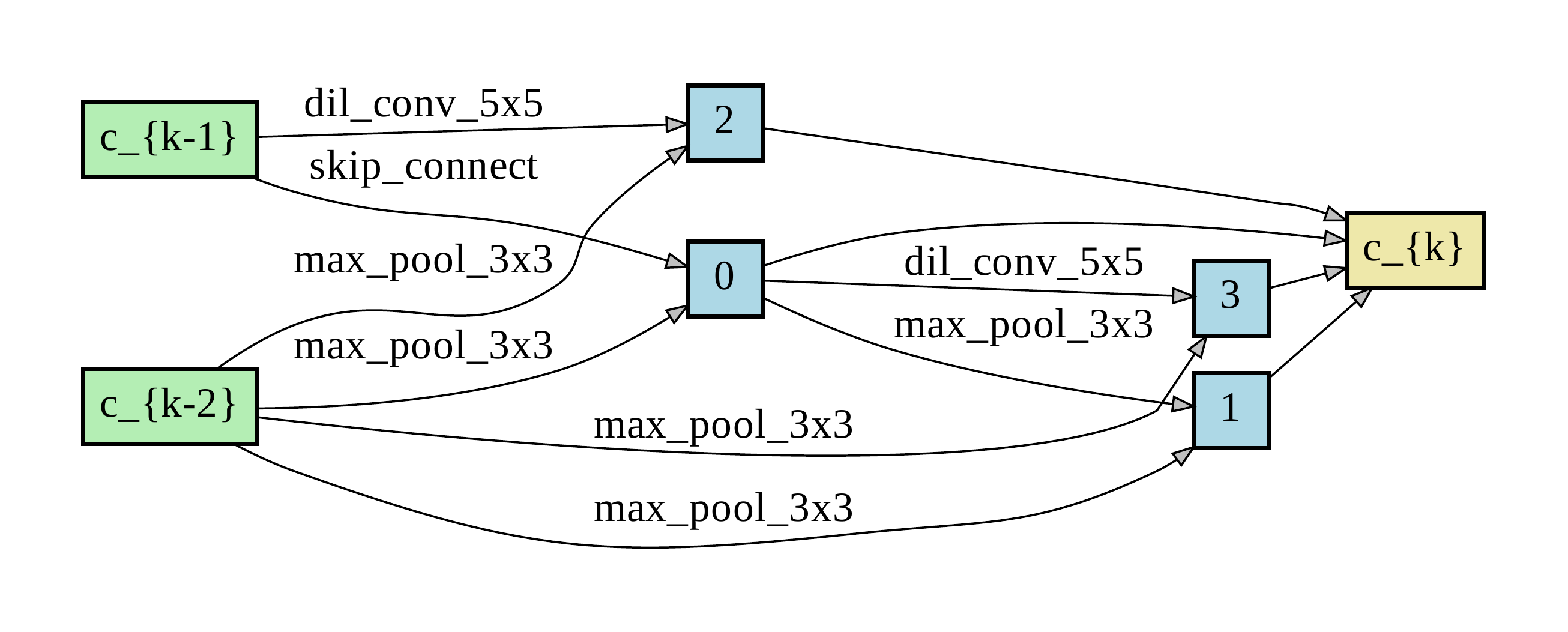}
    \caption{Reduction Cell Architecture}
    \label{fig:reduction}
\end{figure}
We report the architecture searched based on the above non-IID dataset and hyper-parameter setting. Figure \ref{fig:normal} and \ref{fig:reduction} show the normal cell architecture and the reduction cell architecture, respectively. We can see that the reduction cell uses more pooling operations while the normal cell has more convolutional operations.

\section{Future Works}
\label{sec:future_works}
Our future work aims to improve the FedNAS framework from form the following perspectives. 

\begin{itemize} 
\item \textbf{Local NAS under Resource Constraint}. Our current search space fits for cross-organization federated learning, where the edge device can be equipped with powerful GPU devices. But when used in resource-constrained environments such as smartphones or IoT devices, the memory of our search space is too large. Searching on compact search space or using sampling methods are potential solutions to this challenge.

\item \textbf{One Stage NAS}. In our FedNAS framework, we first search the architecture and then train it from scratch. This two-stage training process can be optimized by being merged into a single process. In FedNAS, a naive way to perform one-stage NAS is to continue training the model after the search process is finished. More advanced methods to further improve efficiency are open problems.

\item \textbf{Asynchronous Aggregation}. The searching or training time can be reduced by asynchronous aggregation when each worker has a different number of samples, communication abilities, and computational resources. For FedNAS, the impact of the asynchronous aggregation on model performance should be evaluated.

\item \textbf{Personalized NAS}. Since the data distribution in each client is heterogeneous, a personalized model in each worker may have better accuracy than a global shared model. To meet this requirement, searching for personalized architectures is a challenging task.
\end{itemize}


\end{document}